\documentclass[pdflatex,sn-mathphys-ay,iicol]{sn-jnl}

\newcommand{\benchmark}{NeMoBench}
\newcommand{\task}{NeMo}

\usepackage[table]{xcolor}
\definecolor{my_green}{RGB}{51,102,0}
\definecolor{my_red}{RGB}{204, 0, 0}
\usepackage{pifont}
\newcommand{\cmark}{\textcolor{my_green}{\ding{51}}} 
\newcommand{\xmark}{\textcolor{my_red}{\ding{55}}} 

\usepackage{enumitem}
\setlist[itemize]{noitemsep,leftmargin=*,topsep=0em}
\setlist[enumerate]{noitemsep,leftmargin=*,topsep=0em}
\usepackage{tcolorbox}

\definecolor{my_blue}{HTML}{0000FF}
\definecolor{my_orange}{HTML}{EA7131}

\usepackage{xspace}
\makeatletter
\DeclareRobustCommand\onedot{\futurelet\@let@token\@onedot}
\def\@onedot{\ifx\@let@token.\else.\null\fi\xspace}
\def\eg{\emph{e.g}\onedot} 
\def\ie{\emph{i.e}\onedot}

\makeatother

\usepackage{graphicx}%
\usepackage{multirow}%
\usepackage{amsmath,amssymb,amsfonts}%
\usepackage{amsthm}%
\usepackage[title]{appendix}%
\usepackage{textcomp}%
\usepackage{manyfoot}%
\usepackage{booktabs}%
\usepackage{algorithm}%
\usepackage{algorithmicx}%
\usepackage{algpseudocode}%
\usepackage{listings}%
\usepackage{subcaption}
\usepackage{rotating}
\usepackage{stfloats}
\newcommand{\blfootnote}[1]{%
  \begingroup
    \renewcommand\thefootnote{}%
    \footnotetext{#1}%
  \endgroup
}
\usepackage{pdfpages}

\theoremstyle{thmstyleone}%
\theoremstyle{thmstyletwo}%

\theoremstyle{thmstylethree}%

\begin{document}


\title[Article Title]{
NeMo: \underline{Ne}edle in a \underline{Mo}ntage for Video-Language Understanding
}

\author{
\centering
Zi-Yuan Hu\textsuperscript{*1},
Shuo Liang\textsuperscript{*1},
Duo Zheng\textsuperscript{1},
Yanyang Li\textsuperscript{1}, \\
Yeyao Tao\textsuperscript{1},
Shijia Huang\textsuperscript{1},
Wei Feng\textsuperscript{2},
Jia Qin\textsuperscript{2},
Jianguang Yu\textsuperscript{2}, \\
Jing Huang\textsuperscript{3},
Meng Fang\textsuperscript{4},
Yin Li\textsuperscript{5},
Liwei Wang\textsuperscript{$\dagger$1} \\
\textsuperscript{1}The Chinese University of Hong Kong \quad
\textsuperscript{2}Phoenix TV \quad
\textsuperscript{3}Stanford University \quad
\textsuperscript{4}University of Liverpool \quad
\textsuperscript{5}University of Wisconsin-Madison
}

\abstract{
Recent advances in video large language models (VideoLLMs) call for new evaluation protocols and benchmarks for video-language understanding. Inspired by the needle in a haystack test widely used by LLMs, we introduce a novel task of \textbf{\underline{Ne}edle in a \underline{Mo}ntage} (\textbf{\task{}}), 
{which is designed to assess the temporal understanding capabilities of advanced VideoLLMs.
Specifically, the proposed task focuses on two fundamental abilities critical for temporal understanding, \ie, retrieval-style long-context recall and temporal grounding.
}
To generate video question answering data for our task, we develop a scalable automated data generation pipeline that facilitates high-quality data synthesis. 
Built upon the proposed pipeline, we present \textbf{\benchmark{}}, a video-language benchmark centered on our task. Specifically, our full set of \textit{\benchmark{}} features 31,378 automatically generated question-answer (QA) pairs from 13,486 videos with various durations ranging from seconds to hours. 
Experiments demonstrate that our pipeline can reliably and automatically generate high-quality evaluation data, enabling \textit{\benchmark{}} to be continuously updated with the latest videos.
We evaluate 20 state-of-the-art models on our benchmark, providing extensive results and key insights into their capabilities and limitations. 
Our project page is available at: \url{https://lavi-lab.github.io/NeMoBench}.
}


\maketitle

\blfootnote{$^*$Equal contributions. $^\dagger$Project lead \& corresponding author.}

\begin{figure*}[t!]
\centering
\includegraphics[width=\textwidth]{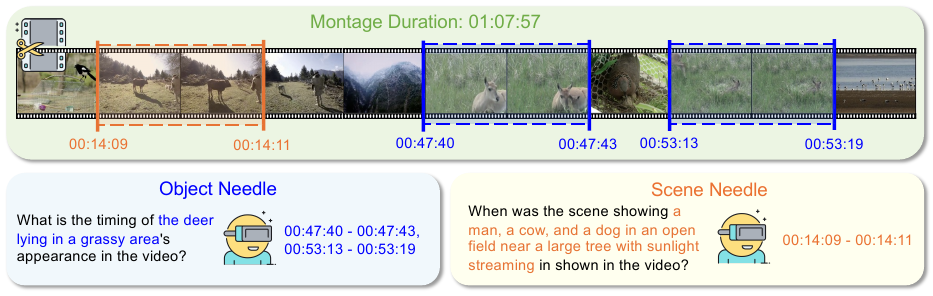}
\caption{
Illustration of our \textbf{\underline{Ne}edle in a \underline{Mo}ntage} (\textbf{\task{}}) task, showcasing examples of object and scene {needles} in an hour-long montage that is synthesized by numerous loosely related short video clips (see more details in Sec.~\ref{sec:task_suite}). 
}
\label{fig:nemo_teaser}
\end{figure*}

\section{Introduction}
Large language models (LLMs)~\citep{DBLP:conf/nips/BrownMRSKDNSSAA20gpt3,DBLP:journals/corr/abs-2302-13971llama,DBLP:journals/corr/abs-2303-08774gpt4} have recently achieved remarkable success in connecting different modalities, such as video and text. This progress has given rise to a new generation of video large language models (VideoLLMs)~\citep{qwen2-vl,DBLP:journals/corr/abs-2409-18938,internvl}, which excel at a wide range of video understanding tasks. There is growing interest in advancing VideoLLMs for long video understanding, {typically defined as comprehending over hour-long video content}~\citep{gemini1.5,shen2024longvu,qwen2.5vl}. Tackling this challenge not only drives innovation in the development of VideoLLMs but also raises important questions about their evaluation. Many existing benchmarks~\citep{jang2017tgif, xu2017video, ssv, yu2019activitynet, xiao2021next, CLEVRER,DBLP:conf/cvpr/BuchEG00N22} {fall short at considering fine-grained temporal cues, making them inadequate for assessing the advanced temporal understanding capabilities needed for comprehending long videos.}

In natural language processing, a widely adopted approach for evaluating long-context recall in LLMs is the \textit{needle in a haystack} test~\citep{nlpneedle,kuratov2024searchneedle,countingstar,ruler}. This method entails embedding a random statement (``needle'') within a long context (``haystack'') and prompting the model to retrieve it. Although highly specialized, this test effectively simulates real-world scenarios in which crucial information is obscured within extensive content windows. Furthermore, studies indicate that performance on this test is predictive of results on more comprehensive benchmarks~\citep{babilong,videohaystack}, {and our empirical results further show that this observation can be extended to the video-language domain (as discussed in Sec.~\ref{sec:exp_analysis}).} 
This methodology has inspired the development of evaluation paradigms for VideoLLMs, as long video understanding in VideoLLMs closely parallels long-context reasoning in conventional LLMs. Several recent studies~\citep{videohaystack,mlvu} have extended needle in a haystack to VideoLLMs.
However, these variants have yet to resolve the fundamental evaluation challenges faced by VideoLLMs. In particular, the injected ``needle'' is often video-irrelevant (\eg, a static image or an unrelated short clip), rather than being embedded within the semantic context of the video ``haystack''. Consequently, existing benchmarks end up addressing a simplified problem that does not fully capture the complexities of video understanding.

To address this problem, we introduce a novel task of {\textbf{\underline{Ne}edle in a \underline{Mo}ntage} (\textbf{\task{}})}, specifically designed for evaluating VideoLLMs, as shown in Fig.\ \ref{fig:nemo_teaser}. 
Rather than injecting content unrelated to the video~\citep{videohaystack,mlvu},
our task embeds one or more video-relevant ``needles'' (\ie, short video clips) within a ``montage'' (\ie, a video composed from many loosely related short video clips). 
In our design, a VideoLLM must retrieve each individual {needle} by identifying its temporal extent. 
{Our task thus assesses two fundamental abilities critical for temporal understanding, namely retrieval-style long-context recall and temporal grounding.}

Although this new task is clearly defined, creating a scalable benchmark remains challenging. Annotating videos, especially those of long duration, demands substantial manual effort, restricting scalability. To overcome this limitation, we propose a comprehensive yet compact video representation and develop a scalable, automated data generation pipeline that synthesizes long montages by seamlessly combining multiple short video clips. Our experiments show that this pipeline effectively generates high-quality data.

Leveraging our automated pipeline, we present \textbf{\benchmark{}}, a new video-language benchmark designed to evaluate the {temporal understanding capabilities of advanced VideoLLMs.} We curate a diverse and up-to-date collection of videos, officially authorized by Phoenix TV\footnote{https://www.phoenixtv.com/.}, a leading media platform serving global audiences. Importantly, the continuous production of high-quality TV programs enables us to regularly update our benchmark with new content, thereby naturally reducing the risk of data contamination in large model evaluations~\citep{nphardeval,c2leva}.

To this end, our benchmark includes two key components: (1) a comprehensive set (\textbf{\benchmark{}-Full}) comprising 31,378 automatically generated question-answer (QA) pairs about objects and scenes from 13,486 videos of varying durations, and (2) a core subset (\textbf{\benchmark{}-Clean}) containing 2,053 manually verified QA pairs from 940 videos, with durations ranging from seconds to hours. 
We also commit to maintaining and expanding the benchmark by incorporating new TV programs over time. 
We anticipate that the benchmark and resources will facilitate the development of video-language learning.

We conduct extensive experiments on the proposed benchmark, evaluating twenty state-of-the-art models, including both open-source VideoLLMs and closed-source multimodal large models. 
Our experimental results demonstrate that the proposed task presents substantial challenges for both open-source and closed-source models, with a significant performance gap compared to human benchmarks. Furthermore, there remains a considerable disparity between the capabilities of open-source and closed-source models. Our benchmark may offer insights for guiding future efforts to advance open-source models toward the performance of closed-source models. We also observe that model rankings remain consistent across both \textit{\benchmark{}-Clean} and \textit{\benchmark{}-Full}, demonstrating that our pipeline can reliably and automatically generate high-quality evaluation data.
{Additionally, experimental results suggest that our \textit{\benchmark{}} can serve as a diagnostic probe that indirectly reflects the video-language understanding capability of advanced VideoLLMs in real-world long videos.}

\begin{sidewaystable*}
\centering
\caption{
Comparisons between our \textbf{\benchmark{}} and recent VideoLLM benchmarks.
Manual/Auto: raw QA pairs are constructed via manual annotation or automated data generation.
Short: less than 2.5 minutes. 
Medium: between 2.5 and 15 minutes.
Long: more than 15 minutes.
Single/Multi: temporal grounding QA pairs with single or multiple targets.
$\spadesuit$: requires manually annotated data from existing datasets.
$\heartsuit$: a subset of tasks requires purely manual annotation.
Official Authorization: \cmark~indicates that all videos are collected with direct authorization from official platforms to ensure long-term usability (see Sec.~\ref{sec:pipeline}); \cmark$^\triangle$~denotes that the authorized videos are derived from Ego4D~\citep{grauman2022ego4d}; \xmark~indicates that the videos are crawled from public platforms (\eg, YouTube and ShutterStock).
}
\resizebox{\textheight}{!}{
\renewcommand{\arraystretch}{1.3}
\begin{tabular}{l|c|c|c|c|ccc|c|cc|c}
\toprule
\multirow{2}{*}{\textbf{Benchmark}} & \multirow{2}{*}{\textbf{Annotator}} & \multirow{2}{*}{\textbf{\#Videos}} & \multirow{2}{*}{\textbf{\#QA Pairs}} & \textbf{Official} & \multicolumn{3}{c|}{\textbf{Video Type}} & \textbf{Temporal} & \multicolumn{2}{c|}{\textbf{\#Targets}}  & \textbf{Evaluation} \\
\cmidrule(lr){6-8} \cmidrule(lr){10-11}
&  & & & \textbf{Authorization} &  \textbf{Short} & \textbf{Medium} & \textbf{Long} & \textbf{Grounding} & \textbf{Single} & \textbf{Multi}  & \textbf{Method} \\
\midrule
\multicolumn{11}{c}{\textit{\textbf{VideoLLM Benchmarks (Manual Annotation)}}} \\ 
\midrule
MVBench~\citep{mvbench}  & Manual$^\spadesuit$ & 3,641 & 4,000 & \xmark & \cmark & \xmark & \xmark & \xmark & \xmark & \xmark   & Rule\\
AutoEval-Video~\citep{autoeval}  & Manual & 327 & 327 & \xmark & \cmark & \xmark & \xmark & \xmark & \xmark & \xmark   & LLM\\
MovieChat-1K~\citep{moviechat}  & Manual & 130 & 1,950 & \xmark & \xmark & \cmark & \xmark & \xmark & \xmark & \xmark   & LLM\\
MMBench-Video~\citep{mmbenchvideo} & Manual & 600 & 1,998 & \xmark & \cmark & \cmark & \xmark & \xmark & \xmark & \xmark  & LLM \\
Seed-Bench (Video)~\citep{seedbench} & Manual$^\heartsuit$ & 3,757 & 3,757 & \xmark & \cmark & \xmark & \xmark & \xmark & \xmark & \xmark  & Rule \\
MMWorld~\citep{mmworld} & Manual$^\heartsuit$ & 1,910 & 6,627 & \xmark & \cmark & \cmark & \cmark & \xmark & \xmark & \xmark  & Rule \\
HourVideo~\citep{chandrasegaran2024hourvideo}  & Manual & 500 & 12,976 & \cmark$^\triangle$ & \cmark & \cmark & \cmark & \xmark & \xmark & \xmark   & Rule\\
LongVideoBench~\citep{longvideobench}  & Manual & 3,763 & 6,678 & \xmark & \cmark & \cmark & \cmark & \xmark & \xmark & \xmark  & Rule\\
LVBench~\citep{lvbench}  & Manual & 103 & 1,549 & \xmark & \xmark & \xmark & \cmark & \xmark & \xmark & \xmark   & Rule\\
Video-MMMU~\citep{videommmu}  & Manual & 300 & 900 & \xmark & \cmark & \cmark & \cmark & \xmark & \xmark & \xmark  & Rule \\
MMVU~\citep{mmvu}  & Manual & 1,529 & 3,000 & \xmark & \cmark & \cmark & \xmark & \xmark & \xmark & \xmark  & Rule/LLM \\
E.T. Bench~\citep{etbench} & Manual$^\spadesuit$ & 7,002 & 7,289 & \xmark & \cmark & \cmark & \xmark & \cmark & \cmark & \cmark   & Rule\\
TVGBench~\citep{timer1} & Manual$^\spadesuit$ & 462 & 800 & \xmark & \cmark & \cmark & \xmark & \cmark & \cmark & \xmark   & Rule\\
MLVU~\citep{mlvu}  & Manual$^\heartsuit$ & 1,730 & 3,102 & \xmark & \xmark & \cmark & \cmark & \xmark & \xmark & \xmark  & Rule/LLM \\
Video-MME~\citep{videomme}  & Manual & 900 & 2,700 & \xmark & \cmark & \cmark & \cmark & \xmark & \xmark & \xmark  & Rule \\
\midrule
\multicolumn{11}{c}{\textit{\textbf{VideoLLM Benchmarks (Automatic Generation)}}} \\ 
\midrule
EgoSchema~\citep{egoschema}  & Auto & 5,063 & 5,063 & \cmark$^\triangle$ & \xmark & \cmark & \xmark & \xmark & \xmark & \xmark   & Rule\\
TempCompass~\citep{tempcompass}  & Auto & 410 & 7,540 & \xmark & \cmark & \xmark & \xmark & \xmark & \xmark & \xmark   & Rule/LLM \\
VideoAutoArena~\citep{VideoAutoArena}  & Auto & 2,881 & - & \xmark & \cmark & \cmark & \cmark & \xmark & \xmark & \xmark  & LLM\\
VNBench~\citep{videohaystack}  & Auto & 1,350 & 1,350 & \xmark & \cmark & \cmark & \xmark & \xmark & \xmark & \xmark  & Rule \\
\midrule
\textbf{\benchmark{}-Full (Ours)} & Auto & \textbf{13,486} & \textbf{31,378} & \cmark & \cmark & \cmark & \cmark & \cmark & \cmark & \cmark   & Rule\\
\textbf{\benchmark{}-Clean (Ours)} & Auto & 940 & 2,053 & \cmark & \cmark & \cmark & \cmark & \cmark & \cmark & \cmark  & Rule \\
\bottomrule
\end{tabular}
}
\label{tab:benchmark_comparison}
\end{sidewaystable*}

\medskip
\textbf{Our main contributions} are threefold:
\begin{itemize}
    \item We propose a novel task, \textit{needle in a montage}, to {evaluate VideoLLMs in video understanding by assessing two fundamental abilities critical for temporal understanding: retrieval-style long-context recall and temporal grounding.}
    \item We introduce a scalable, automated data generation pipeline, enabling the creation of high-quality video-language benchmarks.
    \item We present \textit{\benchmark{}}, a novel benchmark accompanied by extensive experiments, which provide new insights into the strengths and limitations of state-of-the-art models.
\end{itemize}

\section{Related Work}
\subsection{Video Large Language Models}
Many VideoLLMs integrate a pre-trained vision backbone~\citep{clip,siglip,eva-clip} with an LLM through an adapter module~\citep{frozen,Perceiver} that aligns visual and textual data~\citep{DBLP:journals/corr/abs-2409-18938, videollama, maaz2023videochatgpt, mvbench, timechat, tgvid, vtimellm, momentor, tarsier2, aotd,videorag}. 
Early works focus on short-video tasks, while struggling with effectively processing medium and long videos~\citep{videomme, video-llava,longvila}. 
Recent advances in LLMs, especially their improved video understanding capabilities and increased context length, have significantly advanced the development of VideoLLMs~\citep{lin2023vila, zhang2024videoinstructiontuningsynthetic, liu2024oryx, yao2024minicpm, qwen2-vl, gpt4o, qwen2.5vl,timer1}. 
In particular, recent VideoLLMs, including both open-source and closed-source models, are capable of understanding over hour-long videos~\citep{gemini1.5,shen2024longvu}.
However, our experimental results reveal that even the latest VideoLLMs exhibit a significant performance gap when compared to humans on our \textit{\benchmark{}}, underscoring the critical challenges in \textit{\task{}} task.

\begin{figure*}[htbp]
  \centering
  \begin{subfigure}[b]{0.495\textwidth}
    \includegraphics[width=\linewidth]{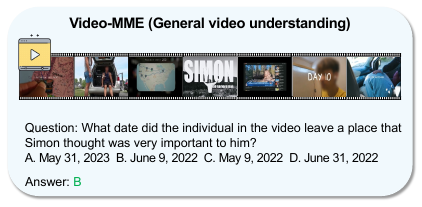}
  \end{subfigure}
  \begin{subfigure}[b]{0.495\textwidth}
    \includegraphics[width=\linewidth]{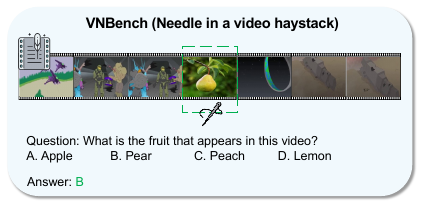}
  \end{subfigure}
  \begin{subfigure}[b]{0.5\textwidth}
    \includegraphics[width=\linewidth]{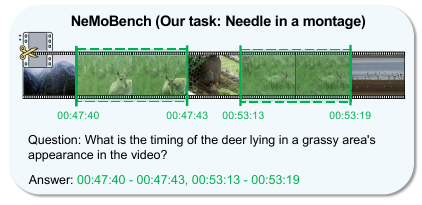}
  \end{subfigure}
  \caption{Comparisons between our \textbf{\benchmark{}} (Sec.~\ref{sec:benchmark}) and other widely-used VideoLLM benchmarks~\citep{videomme,videohaystack}. 
  Our benchmark is centered on the proposed \textit{\task{}} task, {specifically designed to assess two fundamental abilities critical for temporal understanding, namely retrieval-style long-context recall and temporal grounding}.
}
  \label{fig:benchmark_comparison}
\end{figure*}

\subsection{Video-Language Benchmarks}
Numerous benchmarks have been developed to evaluate VideoLLMs across various aspects of video understanding~\citep{autoeval, seedbench, videobench, vlmeval, egoschema, mmbenchvideo, PerceptionTest, rextime, videommmu, mmvu}. 
For example, MVBench~\citep{mvbench}, E.T. Bench~\citep{etbench}, and TVGBench~\citep{timer1} evaluate temporal perception in video understanding, but they are limited to second- and minute-long videos, and rely on manually annotated data from conventional datasets~\citep{gao2017tall,lei2021qv,grauman2022ego4d,DBLP:conf/cvpr/HeilbronEGN15ActivityNet,DBLP:conf/cvpr/Zala0K0OMB23HiREST,DBLP:journals/tacl/RegneriRWTSP13TaCoS}.

Recently, the community has witnessed the emergence of benchmarks for long video understanding~\citep{moviechat, mlvu, mmworld, lvbench, lvhaystack, cgbench, vrbench, temporalbench}.
As shown in Tab.~\ref{tab:benchmark_comparison}, HourVideo~\citep{chandrasegaran2024hourvideo} provides a comprehensive and challenging hour-long benchmark, featuring high-quality QA pairs for egocentric videos~\citep{grauman2022ego4d}.
LongVideoBench~\citep{longvideobench} emphasizes long-context interleaved video-language understanding, while Video-MMMU~\citep{videommmu} and MMVU~\citep{mmvu} target multi-disciplinary video understanding.
Video-MME~\citep{videomme} provides a comprehensive evaluation of video understanding in terms of task diversity and video duration, while LVBench~\citep{lvbench} focuses on extremely long video understanding.
However, these benchmarks still heavily rely on manual annotation, which is both labor-intensive and time-consuming, thereby hindering the scalability of benchmark construction.
Different from prior works, our work makes two key innovations.
First, our benchmark is built on a \textbf{novel task of ``needle in a montage''}, specifically designed to {two fundamental abilities critical for temporal understanding, \ie, retrieval-style long-context recall and temporal grounding.} 
Second, our benchmark is \textbf{automatically constructed} using vision foundation models, covering high-quality test data with high efficiency.

\medskip
\noindent \textbf{VideoLLM benchmarks with ``needle in a haystack''.}
The most relevant works are those considering ``needle in a haystack'' tests for VideoLLM benchmarks~\citep{videohaystack, mlvu}. 
As shown in Fig.~\ref{fig:benchmark_comparison}, VNBench~\citep{videohaystack} involves multiple-choice QA pairs for its ``needle in a video haystack'' task, where text or image needles (\ie, \textit{static needles}) are inserted into a video haystack based on predefined rules. However, since these \textit{static needles} are unrelated to the video (\eg, using a name word or a fruit image), this task fails to effectively challenge advanced VideoLLMs. For example, Gemini-1.5-Pro-002 exhibits superior performance across varying video durations. 
MLVU~\citep{mlvu} also introduces a multiple-choice needle QA task, where a short video needle (\ie, \textit{dynamic needle}) is embedded into a video haystack. 
However, its design of using a video needle distinct from the video haystack (\eg, with different video categories or shooting styles) largely simplifies the question difficulty. 
Additionally, MLVU only considers a single needle per video haystack. 
In contrast, our benchmark is \textbf{more challenging}, as it composes the haystack using loosely related clips and allows for multiple video-relevant \textit{dynamic needles} within the same video haystack, providing a more rigorous evaluation for video understanding.
Furthermore, our task adopts a \textbf{novel task format} of open-ended needle grounding within a montage, rather than multiple-choice needle QA.
In addition, our evaluation benchmark is established with a diverse collection of \textbf{authorized videos} from high-quality TV programs to ensure long-term usability and allow for future expansion. 

\medskip
\noindent \textbf{Automated data generation in VideoLLM benchmarks}.
A growing body of work is exploring automated approaches to constructing video-language benchmarks~\citep{egoschema,tempcompass,videovista,VideoAutoArena}. 
EgoSchema~\citep{egoschema} and TempCompass~\citep{tempcompass} focus on second- and minute-long video understanding, where prompt design plays a central role in prompting LLMs to generate QA pairs from raw videos.
In contrast, our benchmark covers videos with \textbf{diverse durations}, ranging from seconds to hours.
Moreover, we introduce a \textbf{new automated data generation pipeline}, leveraging our proposed \textbf{comprehensive yet compact video representation}.
VideoAutoArena~\citep{VideoAutoArena} introduces an arena-style benchmark~\citep{arena} that employs visual prompting for user simulation to automatically generate open-ended questions from LongVideoBench's videos. 
It largely relies on LLMs to measure model performance in battles, yet such a judging system may exhibit biased scoring, as reported in their paper.
In comparison, our benchmark employs rule-based evaluation metrics, eliminating the need for LLM assistance.

{
\medskip
\noindent \textbf{Temporal grounding benchmarks}.
Temporal grounding benchmarks~\citep{gao2017tall,lei2021qv,grauman2022ego4d,rextime,etbench,momentseeker} represent an important line of research that is relevant to our work.
Charades-STA~\citep{gao2017tall} and QVHighlights~\citep{lei2021qv} represent early efforts in the temporal grounding task. However, they are limited to short videos.
Ego4D-NLQ~\citep{grauman2022ego4d} broadens this task to longer egocentric videos, while ReXTime~\citep{rextime} extends it by integrating temporal grounding with reasoning-based multiple-choice question answering.
E.T. Bench~\citep{etbench} incorporates the temporal grounding task as part of its comprehensive time-sensitive evaluation, while MomentSeeker~\citep{momentseeker} further enhances the task by supporting various video durations and diverse query types.
Despite these advances, these recent efforts still heavily rely on manual annotation to generate QA data (or require manually annotated data from existing datasets), which is both labor-intensive and time-consuming, thereby limiting the scalability and cost-effectiveness of benchmark construction.
Our work differs from prior benchmarks in two major aspects. First, our benchmark is built on a novel task scenario of “needle in a montage”, which enhances and extends the standard “needle in a haystack'' task in terms of needle design (\eg, video-relevant needles, different needle types, varying number of needles), montage design (\eg, montage duration, montage composition), and QA format design (\ie, with the same output format as the conventional temporal grounding task).
Second, our benchmark is automatically constructed using our scalable, automated data generation pipeline, covering high-quality test data with high efficiency.
}

\section{Benchmark Design and Construction}

\subsection{Our Task: Needle in a Montage}
\label{sec:task_suite}
We start by introducing our task of \textbf{{``needle in a montage'' (\task{})}}, tailored for {evaluating advanced VideoLLMs in video-language understanding by assessing two fundamental abilities critical for temporal understanding: retrieval-style long-context recall and temporal grounding.} 
As illustrated in Fig.~\ref{fig:nemo_teaser}, \task{} shares a similar definition as video question answering, where each testing sample consists of
(1) target ``needles'': single or multiple video-relevant short clips inserted into the video montage, each target needle contains distinct video content, yet all refer to the same needle grounding question; (2) ``montage'': a video synthesized by many loosely related video clips from the same video source; (3) a natural language ``needle grounding question'' describing the target needles; and (4) an open-ended ``answer'' identifying the temporal extents of the target needles within the montage.
We now describe our designs of the needle, montage, and QA format, respectively.

\medskip
\noindent\textbf{Needle design.} 
Our needle design encompasses two key aspects, needle type and the number of target needles, which respectively reflect the varying granularity levels and difficulty levels of QA pairs in our task.
\begin{itemize}

\item \textbf{Needle type:}
We define two needle types for our task: \textit{scene} needle and \textit{object} needle. 
The \textit{scene} contains multiple objects and even a broad combination of objects and their interactions, while the \textit{object} refers to a specific instance.
As shown in Fig.~\ref{fig:nemo_teaser}, our task thus comprises two sub-tasks: \textit{scene needle grounding QA} and \textit{object needle grounding QA}.

\item \textbf{Number of target needles:} We consider varying numbers of target needles within the montage.
As in Fig.~\ref{fig:case_num_needle}, this task setting gives rise to two types of QA pairs with different difficulty levels: \textit{single-needle QA} and \textit{multi-needle QA}, with the latter being more challenging, as validated by experimental results.
\end{itemize}

\medskip
\noindent\textbf{Montage design.} To further enhance the diversity and challenges of our task, our montage design consists of two critical aspects: montage duration and montage composition. 
\begin{itemize}
\item \textbf{Montage duration:} We categorize montage durations into three groups: short montage (less than 2.5 minutes), medium montage (between 2.5 and 15 minutes), and long montage (more than 15 minutes).
As shown in Fig.~\ref{fig:case_duration}, this categorization yields three types of QA pairs with increasing difficulty: \textit{short-montage QA}, \textit{medium-montage QA}, and \textit{long-montage QA}. 
This setup is designed to {challenge models' retrieval-style long-context recall and temporal grounding abilities} across varying montage durations.

\begin{figure*}[ht]
\centering
\includegraphics[width=\textwidth]{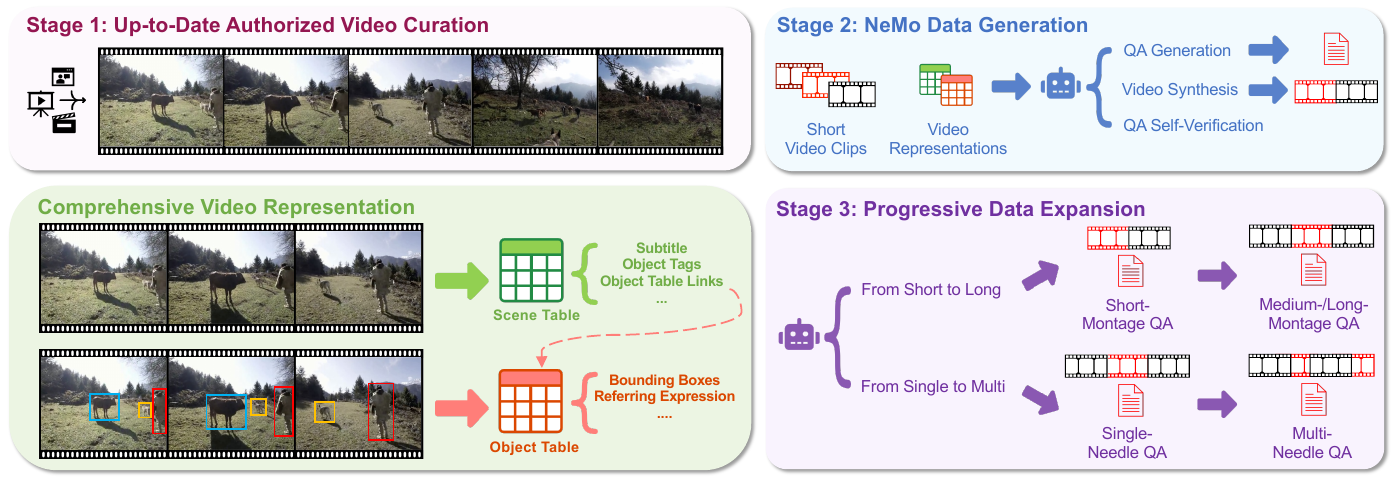}
\caption{Illustrations of our proposed \textbf{comprehensive video representation} and \textbf{automated data generation pipeline}.
Our video representation (Sec.~\ref{sec:video_representation}) consists of scene and object tables, providing critical contextual information for automated data generation.
Our pipeline (Sec.~\ref{sec:pipeline}) facilitates a scalable, cost-effective benchmark construction.
}
\label{fig:overview}
\end{figure*}

\item \textbf{Montage composition:} Unlike~\citep{videohaystack}, which inserts video-irrelevant needles into the video haystack, a montage of our task is constructed by temporally concatenating the target needles with numerous loosely related short video clips. 
Specifically, these video clips are sampled from the target needles' original video source and serve as ``negative'' distractors, forming a montage (\ie, resembling the raw video) with a consistent shooting style. 
We provide the method details in Sec.~\ref{sec:data_gen}.

{Therefore, the target ``needles'' in our task can be considered ``video-relevant needles'', as the target ``needles'' share similar shooting styles, entities, and backgrounds with the surrounding montage clips, while still maintaining a coherent theme across the montage. 
Our visualizations in Sec.~\ref{sec:visualization} further support that the surrounding montage clips share similar visual semantics to the target needles.
Overall, the use of synthetic montages enables cost-effective, scalable benchmark construction, while maintaining high-quality data generation, as discussed in Appendix~\ref{sec:appendix_montage_vs_real}.
}

\end{itemize}

\medskip
\noindent\textbf{QA format design.} 
{We design an open-ended needle grounding QA format following the conventional temporal grounding works}~\citep{gao2017tall,lei2021qv}, where the model is required to output the temporal extents of the target needles within the montage. 
This design facilitates an automatic assessment of the model's performance.

\subsection{Our Automated Data Generation Pipeline}
\label{sec:data_gen}
Rather than relying on manual video annotation, we propose an automated data generation pipeline to create high-quality test data for our task. Generating such data directly from raw videos is particularly challenging, especially when dealing with hour-long videos. To address this challenge, we first propose a video representation containing comprehensive scene- and object-level features, and then introduce an automated pipeline that effectively leverages extracted representations to facilitate high-quality QA generation for our task.

\subsubsection{Comprehensive Video Representation}
\label{sec:video_representation}
Given the inherent spatial and temporal redundancy in videos, we propose a comprehensive yet compact video representation.
Specifically, it follows a hierarchical structure enriched with semantic features, incorporating scene-level and object-level features (see Fig.~\ref{fig:overview}). 
To efficiently handle long videos, we first segment the raw video into multiple short video clips, and then extract the proposed video representation for each clip.
Motivated by~\citep{DBLP:conf/emnlp/ZhongHL024}, we present these video representations in a tabular format to facilitate subsequent automated data generation (see more details in Appendix~\ref{sec:appendix_pipeline}):
\begin{itemize}
\item \textbf{Scene table} encompasses scene-level coarse-grained information of a short video clip, including its temporal extent (\ie, start and end timestamps) in the raw video, subtitles, object tags, and links to all associated object tables within this video clip. 
\item \textbf{Object table} comprises object-level fine-grained information of a specific object, including its temporal extent within the scene, object tag, referring expression, and tracking sequence (\ie, bounding boxes). 
\end{itemize}

\begin{table*}[ht]
\centering
\caption{The statistics of \textit{\benchmark{}-Clean}. More details are provided in Appendix~\ref{sec:appendix_benchmark_statistics}.}
\resizebox{\textwidth}{!}{
\begin{tabular}{c|cccccccccc}
\toprule
\multicolumn{1}{c|}{\multirow{2}{*}{\textbf{Montage Type}}}  & \multicolumn{1}{c}{\multirow{2}{*}{\textbf{\#Needles}}}  & \multicolumn{3}{c}{\textbf{\#Needle Grounding QA Pairs}} & \multicolumn{3}{c}{\textbf{Montage Duration}} & \textbf{Avg \#Needles} & \textbf{Avg Needle} \\
\cmidrule(lr){3-5} \cmidrule(lr){6-8}
& & \textbf{Object} & \textbf{Scene} & \textbf{All} & \textbf{Max} & \textbf{Avg} & \textbf{Min} & \textbf{Per Question} & \textbf{Duration}\\
\midrule

\textbf{Short Montage} & Single-Needle   & 419 & 431 & \multirow{2}{*}{1,246} & \multirow{2}{*}{83.48 s} & \multirow{2}{*}{26.20 s} & \multirow{2}{*}{10.56 s} & \multirow{2}{*}{3.19} & \multirow{2}{*}{3.27 s} \\
 ($<$ 2.5 min)                     & Multi-Needle & 205 & 191 & \\
\midrule
{\textbf{Medium Montage}}& Single-Needle   & 159 & 175 & \multirow{2}{*}{507} & \multirow{2}{*}{13.62 min} & \multirow{2}{*}{5.13 min} & \multirow{2}{*}{2.51 min} & \multirow{2}{*}{3.33} & \multirow{2}{*}{3.28 s}\\
    (2.5 - 15 min)                   & Multi-Needle & 94  & 79  \\
\midrule
{\textbf{Long Montage} }  & Single-Needle   & 98  & 99 & \multirow{2}{*}{300} & \multirow{2}{*}{67.96 min} & \multirow{2}{*}{29.68 min} & \multirow{2}{*}{15.50 min} & \multirow{2}{*}{3.25} & \multirow{2}{*}{3.33 s} \\
     ($>$ 15 min)                  & Multi-Needle & 53  & 50  \\
\midrule
\textbf{All}  & All   & 1,028  & 1,025 & 2,053 & - & -& -& -& - \\       
\bottomrule
\end{tabular}
}
\label{tab:benchmark_statistics}
\end{table*}

\subsubsection{Our Pipeline Design}
\label{sec:pipeline}
In practice, high-quality videos (\eg, television programs) are typically produced through a meticulous video editing process, which involves selecting and editing appropriate video clips from a large-scale repository of footage, seamlessly integrating them into a coherent sequence, and performing rigorous verification for quality control. 
Inspired by this production process, we propose an automated data generation pipeline that follows a similar workflow, thereby facilitating the synthesis of long montages.
As shown in Fig.\ \ref{fig:overview}, our pipeline consists of three stages to create high-quality testing QA pairs for our task (see more details in Appendix~\ref{sec:appendix_pipeline}).

\medskip
\noindent\textbf{Stage 1: Up-to-date authorized video curation.}
We curate a diverse set of up-to-date, high-quality videos from Phoenix TV, a global media platform, with their official authorization.
Moreover, we rigorously ensure that these videos are not included in widely used video-language training datasets~\citep{webvid,mvbench,sharegpt4video,zhang2024videoinstructiontuningsynthetic,llavaonevision,panda70m,HD-VILA-100M,internvid,maaz2023videochatgpt,videochat,LanguageBind}, thereby preventing potential data contamination in large model evaluations.
Subsequently, we extract the comprehensive video representations of these videos to facilitate automated data generation.

\medskip
\noindent\textbf{Stage 2: NeMo data generation.}
In this stage, our pipeline decomposes the data generation into a sequence of sub-tasks:
\begin{enumerate}[noitemsep, nolistsep]
\item {Based on the extracted video representations, select a scene from the raw videos as the \textit{target scene}, and identify a prominent object within it as the \textit{target object}.} Employ visual prompting with a frozen multimodal model (\ie, GPT-4o~\citep{gpt4o}) to generate needle grounding questions for both the \textit{target scene} and \textit{target object}, with their corresponding temporal extents serving as the ground-truth answers.

\item {Sample ``negative'' scenes} from the same video source to preserve a consistent shooting style, ensuring their extracted scene tables contain no object tag corresponding to the \textit{target object}. Compose a long montage by temporally concatenating the \textit{target scene} with these loosely related, ``negative'' scenes.

\item Employ the same frozen multimodal model to perform self-verification to filter out low-quality generated QA pairs, mitigating potential inaccuracies and hallucinations introduced by annotation models.
\end{enumerate}

\medskip
\noindent\textbf{Stage 3: Progressive data expansion.}
Built upon the data generated in Stage 2, we progressively conduct data expansion as follows:
\begin{enumerate}[noitemsep, nolistsep]
\item \textbf{From short to long}: For \textit{short-montage QA} pairs, we extend their montage durations to create \textit{medium-montage} and \textit{long-montage QA} pairs. 
Similar to Stage 2, we synthesize longer montages by seamlessly integrating the \textit{target scene} with more ``negative'' scenes. 

\item \textbf{From single to multi}: For \textit{single-needle QA} pairs, we further expand the number of target needles to create more \textit{multi-needle QA} pairs. Specifically, we divide the \textit{target scene} into multiple segments and reconstruct a montage with multiple target needles.
\end{enumerate}

\subsection{Our Benchmark: \benchmark{}}
\label{sec:benchmark}
Built upon the proposed scalable, automated data generation pipeline, we establish \textbf{\benchmark{}}, a new video-language benchmark centered on our \textit{\task{}} task, which is specifically designed to probe into VideoLLMs' {temporal understanding} capabilities, including {retrieval-style long-context recall and temporal grounding}. 
Specifically, our benchmark features 
(1) \textbf{\benchmark{}-Full}: a full set of 31,378 automatically generated QA pairs from 13,486 videos;
(2) \textbf{\benchmark{}-Clean}: a core set of 2,053 QA pairs sampled from \textit{\benchmark{}-Full} (see Tab.~\ref{tab:benchmark_statistics}) and manually verified to facilitate a robust, precise evaluation of VideoLLMs (see more details in Appendix~\ref{sec:appendix_pipeline}).

Tab.~\ref{tab:benchmark_comparison} compares our \textit{\benchmark{}} with other VideoLLM benchmarks. 
Specifically, \textit{\benchmark{}} is developed using up-to-date authorized videos to ensure long-term usability and mitigate the risk of data leakage.
Our automatically generated benchmark contains at least 2.4$\times$ more QA pairs than other benchmarks, while maintaining data quality.
Moreover, our benchmark covers various task settings (\eg, needle types, montage durations, number of target needles), and spans diverse videos with many scene and object categories (see further details in Appendix~\ref{sec:appendix_benchmark_statistics}).

\begin{table*}[ht]
\centering
\caption{Data quality comparisons between \textit{ManualAnno} and \textit{NeMo-Auto} (Sec.~\ref{sec:gen_data_quality_comparison}).}
\resizebox{0.6\textwidth}{!}{
\begin{tabular}{l|c|cc}
\toprule
\multirow{2}{*}{\textbf{Method}}  & \textbf{Questions}  & \multicolumn{2}{c}{\textbf{Target Needles}} \\
\cmidrule{2-4}
& \textbf{Accuracy} & \textbf{Recall@1x, tIoU=0.7} & \textbf{Average mAP}\\
\midrule

\textit{ManualAnno} & 91.00   & 95.83 & 99.33 \\
\textit{NeMo-Auto}  & 82.50 & 86.65 & 97.59 \\
\bottomrule
\end{tabular}
}
\label{tab:gen_data_compare}
\end{table*}

\section{{Analysis of the Automated Data Generation Pipeline}}
In this section, we present a comparative analysis of our {data generation pipeline} in terms of time efficiency and data quality, incorporating both theoretical and practical perspectives.

\subsection{Time Efficiency Comparison}
\label{sec:gen_data_cost_comparison}
As described in Sec.~\ref{sec:benchmark}, the raw QA pairs of \textit{\benchmark{}-Full} and \textit{\benchmark{}-Clean} are both automatically generated, with the latter involving a minor amount of manual effort for QA verification to ensure a precise VideoLLM benchmarking.
Therefore, to analyze the time efficiency of our {data generation pipeline}, we compare the manual time cost of the conventional manual annotation process~\citep{lei2021qv} (termed \textit{ManualAnno}) with that of our automated data generation pipeline (termed \textit{NeMo-Auto}).

\medskip
\noindent\textbf{Theoretical cost reduction.}
We provide the details of the theoretical analysis in Appendix~\ref{sec:appendix_gen_data_compare}. 
Specifically, \textit{ManualAnno} consists of three manual steps during its annotation process (\ie, \textit{Question construction phase}, \textit{Answer construction phase}, and \textit{Cleaning phase}), while \textit{NeMo-Auto} only involves the \textit{Cleaning phase}.
As discussed in the theoretical analysis, \textit{NeMo-Auto} exhibits superior time efficiency compared with \textit{ManualAnno} across different video durations. 
To be specific, the time reduction ratio (denoted by $\alpha$) can be approximated as a monotonically increasing function.
For short- and medium-duration videos, we have $\alpha_\text{short} \approx \alpha_\text{medium} > 0.8$. 
For long-duration videos, we have $\alpha_\text{long} \approx 0.67$. Therefore, the average time efficiency is computed as $\alpha_\text{average} = (\alpha_\text{short}+\alpha_\text{medium}+\alpha_\text{long}) / 3 >$ \textbf{0.76}, suggesting more than 4x reduction in time. 

\medskip
\noindent\textbf{Practical cost reduction.}
To present the practical time efficiency of \textit{NeMo-Auto}, we recruit seven human annotators to conduct a comparison experiment.
To be specific, the annotators are required to construct 144 QA pairs, 
distributed equally across short-, medium\nobreakdash-, and long-duration videos (\ie, 48 QA pairs per video type). 
The results reveal that, while \textit{ManualAnno} costs 15.9 hours during its annotation, \textit{NeMo-Auto} only requires 3.5 hours for QA verification, yielding a time reduction ratio of $1-\frac{3.5}{15.9}\approx$ \textbf{0.78}, \ie, $\sim$4.5x reduction in time. 
These results corroborate the theoretical analysis, highlighting the significant time efficiency of \textit{NeMo-Auto}.

\subsection{Data Quality Comparison}
\label{sec:gen_data_quality_comparison}
We also attach importance to the data quality of our {data generation pipeline}.
As detailed in Appendix~\ref{sec:appendix_gen_data_compare}, both \textit{ManualAnno} and \textit{NeMo-Auto} conduct the \textit{Cleaning phase} after generating the raw QA pairs. 
Specifically, \textit{ManualAnno}'s raw QA pairs are manually constructed via \textit{Question construction phase} and \textit{Answer construction phase}, while \textit{NeMo-Auto}'s raw QA pairs are automatically generated using our data generation pipeline.
Based on the cleaning results in the \textit{Cleaning phase}, we report both the question quality (\ie, the ratio of raw questions without the need for refining), and answer quality (\ie, Recall@1x, tIoU=0.7 and Average mAP metrics between the raw answers and the refined ones) in Tab.~\ref{tab:gen_data_compare}. 
{The results demonstrate that our \textit{NeMo-Auto} can automatically generate near-human, high-quality evaluation data.}

\section{Experiments and Results}
\label{sec:experimental_results}

\subsection{Experimental Setup}
\label{sec:exp_setup}

\noindent\textbf{Baselines.}
To provide a comprehensive analysis of existing models on our novel task, we evaluate 20 advanced models, covering 2 non-generative specialist models for temporal grounding, 13 open-source VideoLLMs, and 5 closed-source VideoLLMs.
Specifically, the non-generative specialist models include Moment-DETR~\citep{lei2021qv} and UniVTG~\citep{univtg}. 
The open-source VideoLLMs comprise 
LLaVA-Video-7B, LLaVA-Video-72B~\citep{zhang2024videoinstructiontuningsynthetic},
LongVA-7B~\citep{zhang2024longva},
VILA1.5-13B, VILA1.5-40B~\citep{lin2023vila},
Oryx-34B~\citep{liu2024oryx},
MiniCPM v2.6-8B~\citep{yao2024minicpm},
LongVU-7B~\citep{shen2024longvu},
Qwen2.5-VL-7B, Qwen2.5-VL-72B~\citep{qwen2.5vl},
E.T. Chat-3.8B~\citep{etbench},
Time-R1-3B, and Time-R1-7B~\citep{timer1},
with the latter five being specialist VideoLLMs for temporal grounding\footnote{These open-source VideoLLMs involve training data specifically designed for temporal grounding.}.
The closed-source VideoLLMs include Qwen-VL-Max~\citep{Qwen-VL}, GPT-4o~\citep{gpt4o}, and Gemini-series models~\citep{gemini1.5} (\ie, Gemini-1.5-Flash-8B, Gemini-1.5-Flash-002, and Gemini-1.5-Pro-002).

\medskip
\noindent\textbf{Human performance.}
To ensure a comprehensive evaluation, we also include human performance on our \textit{\benchmark{}}.
Given the high cost of human evaluation, we sample 10\% of the QA pairs from our benchmark to assess the human performance. 
Moreover, we ensure that each question is answered by three different human experts, none of whom have previously seen the corresponding videos, thereby mitigating the risk of data leakage.

\begin{figure}[t!]
\begin{center}
\begin{tcolorbox}[colback=black!5!white, colframe=black!75!black, title=\textit{video first}, width=0.95\linewidth]
\textcolor{blue}{\{frame\_0\} \{frame\_1\} ...} \\
There are \textcolor{blue}{\{total\_frames\}} frames from a video with a frame rate of \textcolor{blue}{\{fps\}} FPS. The frames were taken at the following times: 0.0s, 1.0s, ... \\
Carefully watch the video and answer the question. \\
Question: \textcolor{blue}{\{question\}} \\
Note: \\
1. Your response should be a single line. Each line follows the format as ``Answer:  {[start\_time]} -  {[end\_time]}'', where the  {[start\_time]} 
and  {[end\_time]} mean the start time and end time of the object or 
scene in this video, respectively. \\
2. There may be single or multiple possible time intervals. 
Please list them all by following the previous time format, separated by ``,''.
\end{tcolorbox}
\begin{tcolorbox}[colback=black!5!white, colframe=black!75!black, title=\textit{interleaved with timestamp}, width=0.95\linewidth]
There are \textcolor{blue}{\{total\_frames\}} frames from a video with a frame rate of \textcolor{blue}{\{fps\}} FPS. \\
Frame 0 (sampled at 0.0s) \textcolor{blue}{\{frame\_0\}} \\
Frame 1 (sampled at 1.0s) \textcolor{blue}{\{frame\_1\}} \\
... \\
Carefully watch the video and answer the question. \\
Question: \textcolor{blue}{\{question\}} \\
Note: \\
...
\end{tcolorbox}
\caption{Prompt design on \textit{\benchmark{}}.}
\label{fig:prompt_design}
\end{center}
\end{figure}

\medskip
\noindent\textbf{Sampling strategy.}
For model input, video frames are sampled at a rate of 1 FPS.
If the total number of frames exceeds a predetermined maximum frame count $N$, which is mainly determined
by the model's default configuration from~\citep{videomme}, a uniform sampling strategy is applied to retain $N$ frames.
Specifically, for open-source models, $N$ is set to 
64 for LLaVA-Video-7B, LLaVA-Video-72B, and MiniCPM v2.6-8B, 
128 for LongVA-7B and Oryx-34B, 
768 for Qwen2.5-VL-7B, Qwen2.5-VL-72B, Time-R1-3B, and Time-R1-7B,
1024 for VILA1.5-13B and VILA1.5-40B,
and 3600 for LongVU-7B and E.T. Chat-3.8B.
For closed-source models, $N$ is set to 
160 for GPT-4o, 
3600 for Gemini-series models, 
and 350/350/330 for Qwen-VL-Max on short/medium/long montages, respectively.

\medskip
\noindent\textbf{Prompt design.}
For a fair comparison, we choose the best-performing prompt for each VideoLLM during evaluation. 
Specifically, most open-source models utilize the \textit{video first} prompt (see Fig.~\ref{fig:prompt_design}) to match their default input format, while some models (\ie, Qwen2.5-VL and Time-R1) employ their customized prompts (see Appendix~\ref{sec:appendix_prompt_design}).
For closed-source models, experimental results suggest that the \textit{interleaved with timestamp} prompt is more effective.

\medskip
\noindent\textbf{Evaluation metrics.}
Inspired by~\citep{grauman2022ego4d}, we use Recall@1x, tIoU=0.7 (used as the primary metric for experimental analysis), Recall@1x, tIoU=0.5, and Average mAP as the automatic evaluation metrics to facilitate a comprehensive assessment.
More details of the evaluation are provided in Appendix~\ref{sec:appendix_setup}, including the evaluation protocol, the calculation of tIoU, and the refusal rates of closed-source models.

\subsection{Benchmark Results}
\label{sec:main_result}

\noindent\textbf{Open-source models struggle on the proposed task.}
Experimental results reveal substantial limitations of existing open-source models on our benchmark.
In Tab.~\ref{tab:short_exp_clean}, we observe that the best-performing VideoLLM (\ie, Qwen2.5-VL-72B) and non-generative specialist model (\ie, UniVTG) only achieve 28.04\% and 13.00\% Recall@1x, tIoU=0.7 on object needle grounding, respectively.
Moreover, Tab.~\ref{tab:mid_exp_clean} presents that these open-source models exhibit a drastic performance drop on longer montages, attributed to their limited context length or weak temporal grounding ability.
Specifically, even the best open-source model exhibits an 18.47\% drop (using the Avg metric) on medium montages compared with the results on short montages. 

\begin{table*}[p]
\centering
\caption{Evaluation results on \textit{\benchmark{}-Clean} (short montages). 
\textbf{Avg}: Average Recall@1x, tIoU=0.7 across object and scene needle grounding.
\textbf{Bold}: Best result.
\underline{Underline}: Second-best result.
\textbf{Rank}: Performance rankings of VideoLLMs.
}
\resizebox{\textwidth}{!}{
\begin{tabular}{l|ccc|ccc|cc}
\toprule
\multirow{3}{*}{\textbf{Methods}} &  \multicolumn{3}{c|}{\textbf{Short Object Needle Grounding}} & \multicolumn{3}{c|}{\textbf{Short Scene Needle Grounding}} & \multirow{3}{*}{\textbf{Avg}} & \multirow{3}{*}{\textbf{Rank}} \\
\cmidrule{2-7}
& \textbf{Recall@1x,} & \textbf{Recall@1x,} & \textbf{Average}  & \textbf{Recall@1x,} & \textbf{Recall@1x,} & \textbf{Average} \\
& \textbf{tIoU=0.7}   & \textbf{tIoU=0.5}   & \textbf{mAP}   & \textbf{tIoU=0.7}   & \textbf{tIoU=0.5}   & \textbf{mAP}   \\ 
\midrule
\multicolumn{9}{c}{\textit{\textbf{Non-Generative Specialist Models}}} \\ 
\midrule
Moment-DETR~\citep{lei2021qv} &  7.09 &  17.17 &  40.78   & 10.38  & 23.19 & 51.20  & 8.74  & - \\ 
UniVTG~\citep{univtg} & 13.00  & 24.70  & 53.16  & 19.65  & 34.42 & 64.19  & 16.33  & - \\ 
\midrule
\multicolumn{9}{c}{\textit{\textbf{Open-Source VideoLLMs}}} \\ 
\midrule
LLaVA-Video-7B~\citep{zhang2024videoinstructiontuningsynthetic}       & 1.10  & 1.90  & 11.27    & 2.34  & 4.07  & 19.48   & 1.72 & 18  \\ 
LongVA-7B~\citep{zhang2024longva}            & 1.20  & 4.49  & 18.19    & 2.54  & 6.51  & 21.87   & 1.87 & 17 \\ 
VILA1.5-13B~\citep{lin2023vila}             & 0.80  & 1.90  & 4.72     & 3.36  & 6.21  & 14.77   & 2.08 & 16 \\ 
Oryx-34B~\citep{liu2024oryx}             & 2.69  & 8.88  & 34.95    & 5.39  & 13.84  & 41.40   & 4.04 & 15 \\ 
MiniCPM v2.6-8B~\citep{yao2024minicpm}      & 4.49  & 10.68 & 36.94    & 3.87  & 12.61 & 41.75   & 4.18 & 14 \\ 
VILA1.5-40B~\citep{lin2023vila}             & 4.49  & 10.88 & 27.50    & 6.82  & 16.68 & 40.84   & 5.66 & 13 \\ 
LLaVA-Video-72B~\citep{zhang2024videoinstructiontuningsynthetic}      & 8.88  & 18.36 & 40.40    & 12.31 & 22.08 & 41.89   & 10.60 & 12 \\ 
E.T. Chat-3.8B~\citep{etbench}            & 9.72 & 21.54 & 38.23    & 14.94 & 26.82 & 45.56 & 12.33 &  11 \\ 
LongVU-7B~\citep{shen2024longvu}            & 15.37 & 31.44 & 63.06    & 19.43 & 34.18 & 68.48   & 17.40 & 10 \\ 
{Time-R1-3B}~\citep{timer1} & 15.27 &32.53 & 63.84&22.38&40.18&72.85&18.82 & 9 \\
{Time-R1-7B}~\citep{timer1} & 22.36 & 38.02 & 68.25 & 29.70 & 44.86 & 74.76 & 26.03 & 8 \\
{Qwen2.5-VL-7B}~\citep{qwen2.5vl} & 24.45 & 32.63 & 57.22 & 29.70 & 41.71 & 65.62 & 27.08 & 7 \\
{Qwen2.5-VL-72B}~\citep{qwen2.5vl} & 28.04 & 37.62 & 61.01 & 31.11 & 43.13 & 66.66 & 29.69 &  6 \\
\midrule
\multicolumn{9}{c}{\textit{\textbf{Closed-Source VideoLLMs}}} \\ \midrule
Gemini-1.5-Flash-8B~\citep{gemini1.5}  & 41.86 & 50.85 & 73.89    & 55.70 & 63.44 & 82.38   & 48.78 & 5 \\ 
Qwen-VL-Max~\citep{Qwen-VL}          & 54.46 & 63.69 & 82.77   & 56.63 & 64.80 & 83.03   & 55.55 & 4 \\ 
Gemini-1.5-Flash-002~\citep{gemini1.5} & 62.64 & 73.73 & 85.24    & \underline{72.10} & \underline{81.26} & \underline{90.15}  & 67.37 & 3 \\ 
GPT-4o~\citep{gpt4o}               & \underline{64.07} & \underline{74.85} & \underline{88.56}    & 71.41 & 81.18 & 89.91   & \underline{67.74} & 2 \\ 
Gemini-1.5-Pro-002~\citep{gemini1.5}   &   \textbf{65.83} &   \textbf{78.22} &   \textbf{89.81}   &  \textbf{74.85} &  \textbf{85.54} &  \textbf{92.85} & \textbf{70.34} & 1 \\ 
\midrule
\multicolumn{9}{c}{\textit{\textbf{Human Performance}}} \\ \midrule
Human Experts   & 93.62 & 96.10 & 95.31     & 94.33 & 95.39 & 96.31  & 93.98 & - \\ 
\bottomrule
\end{tabular}
}
\label{tab:short_exp_clean}
\end{table*}

\begin{table*}[p]
\centering
\caption{Evaluation results on \textit{\benchmark{}-Clean} (medium montages). 
\textbf{Avg}: Average Recall@1x, tIoU=0.7 across object and scene needle grounding.
\textbf{Bold}: Best result.
\underline{Underline}: Second-best result.
\textbf{Rank}: Performance rankings of VideoLLMs.
}
\resizebox{\textwidth}{!}{
\begin{tabular}{l|ccc|ccc|cc}
\toprule

\multirow{3}{*}{\textbf{Methods}} &  \multicolumn{3}{c|}{\textbf{Medium Object Needle Grounding}} & \multicolumn{3}{c|}{\textbf{Medium Scene Needle Grounding}} & \multirow{3}{*}{\textbf{Avg}} & \multirow{3}{*}{\textbf{Rank}} \\
\cmidrule{2-7}
& \textbf{Recall@1x,} & \textbf{Recall@1x,} & \textbf{Average}  & \textbf{Recall@1x,} & \textbf{Recall@1x,} & \textbf{Average} \\
& \textbf{tIoU=0.7}   & \textbf{tIoU=0.5}   & \textbf{mAP}   & \textbf{tIoU=0.7}   & \textbf{tIoU=0.5}   & \textbf{mAP}   \\ 
\midrule
\multicolumn{9}{c}{\textit{\textbf{Open-Source VideoLLMs}}} \\ 
\midrule
MiniCPM v2.6-8B~\citep{yao2024minicpm}      & 0.23  & 0.68  & 4.79      & 0.74  & 2.22  & 6.48  & 0.49 & 13 \\ 
{Time-R1-3B}~\citep{timer1} & 0.45 & 1.36 &4.27 & 0.74 & 2.47 & 5.81 & 0.60 & 12 \\
LongVU-7B~\citep{shen2024longvu}            & 0.68  & 1.82  & 3.56     & 1.23  & 2.22 & 4.05 & 0.96 & 11  \\ 
E.T. Chat-3.8B~\citep{etbench}   & 0.76 & 1.76 & 4.25    & 1.62 & 3.50 & 6.27 &  1.19 & 10  \\ 
LLaVA-Video-72B~\citep{zhang2024videoinstructiontuningsynthetic}      & 1.59  & 2.50  & 6.82   & 0.99  & 2.96  & 8.72  & 1.29 & 9   \\
{Qwen2.5-VL-7B}~\citep{qwen2.5vl} & 5.39 & 7.49 & 21.25 & 8.28 & 13.45 & 29.55 & 6.83 & 8 \\
{Time-R1-7B}~\citep{timer1} & 7.95 & 15.91 & 35.33 &  11.88 & 20.54 & 44.95 & 9.92 & 7 \\
{Qwen2.5-VL-72B}~\citep{qwen2.5vl} &  8.86 & 16.82 & 35.95 & 13.58  & 27.41 & 51.35 & 11.22 & 6 \\
\midrule

\multicolumn{9}{c}{\textit{\textbf{Closed-Source VideoLLMs}}} \\ \midrule
Qwen-VL-Max~\citep{Qwen-VL}          & 12.44  & 16.67  & 34.16 & 17.14 & 21.23 & 40.55 & 14.79 & 5 \\ 
GPT-4o~\citep{gpt4o}               & 24.32  & 38.41  & 63.97   & 28.64  & 43.95  & 69.75 &  26.48 & 4   \\ 
Gemini-1.5-Flash-8B~\citep{gemini1.5}  & 32.71 & 40.71 & 59.00   & 43.81 & 53.35 & 71.91 & 38.26 & 3 \\ 
Gemini-1.5-Flash-002~\citep{gemini1.5} & \underline{41.41} & \underline{52.24} & \underline{69.51}    & \underline{55.41} & \underline{68.30} & \underline{81.85}  & \underline{48.41} &  2 \\ 
Gemini-1.5-Pro-002~\citep{gemini1.5}   &  \textbf{48.24}  &  \textbf{58.35}  &  \textbf{74.30}   &  \textbf{65.46} &  \textbf{75.77}  &  \textbf{86.08}   & \textbf{56.85} & 1  \\ 
\midrule
\multicolumn{9}{c}{\textit{\textbf{Human Performance}}} \\ 
\midrule
Human Experts   & 90.74 & 88.89 & 91.87        & 90.32 & 89.25 & 93.37 & 90.53 & - \\ 
\bottomrule
\end{tabular}
}
\label{tab:mid_exp_clean}
\end{table*}


\begin{table*}[t!]
\centering
\caption{Evaluation results on \textit{\benchmark{}-Clean} (long montages). 
\textbf{Avg}: Average Recall@1x, tIoU=0.7 across object and scene needle grounding.
\textbf{Bold}: Best result.
\underline{Underline}: Second-best result.
\textbf{Rank}: Performance rankings of VideoLLMs.
}
\resizebox{\textwidth}{!}{
\begin{tabular}{l|ccc|ccc|cc}
\toprule

\multirow{3}{*}{\textbf{Methods}} &  \multicolumn{3}{c|}{\textbf{Long Object Needle Grounding}} & \multicolumn{3}{c|}{\textbf{Long Scene Needle Grounding}} & \multirow{3}{*}{\textbf{Avg}} & \multirow{3}{*}{\textbf{Rank}} \\
\cmidrule{2-7}
& \textbf{Recall@1x,} & \textbf{Recall@1x,} & \textbf{Average}  & \textbf{Recall@1x,} & \textbf{Recall@1x,} & \textbf{Average} \\
& \textbf{tIoU=0.7}   & \textbf{tIoU=0.5}   & \textbf{mAP}   & \textbf{tIoU=0.7}   & \textbf{tIoU=0.5}   & \textbf{mAP}   \\

\midrule
\multicolumn{9}{c}{{\textit{\textbf{Open-Source VideoLLMs}}}} \\ 
\midrule
{Time-R1-3B~\citep{timer1}} & {0.00} & {0.00} & {0.20} & {0.00} & {0.00} & {0.16} & {0.00} & {9} \\
{Time-R1-7B~\citep{timer1}} & {0.81} & {1.62} & {5.23} & {0.00} & {1.24} & {4.50} & {0.40} & {8} \\
{Qwen2.5-VL-7B~\citep{qwen2.5vl}} & {0.40} & {0.81} & {1.92} & {0.41} & {0.83} & {1.41} & {0.41} & {6} \\
{Qwen2.5-VL-72B~\citep{qwen2.5vl}} & {0.41} & {0.81} & {1.50} & {0.41} & {1.24} & {2.68} & {0.41} & {6} \\
\midrule

\multicolumn{9}{c}{\textit{\textbf{Closed-Source VideoLLMs}}} \\ \midrule
Qwen-VL-Max~\citep{Qwen-VL}          & 0.94  & 1.88  & 4.83   & 0.95  & 3.32  & 7.12  & 0.95 & 5 \\ 
GPT-4o~\citep{gpt4o}               & 2.83  & 7.29  & 23.20     & 2.90  & 7.88  & 26.74 & 2.87 & 4 \\ 
Gemini-1.5-Flash-8B~\citep{gemini1.5}  & 25.53 & 34.04 & 47.63  & 39.30 & 44.54 & 60.97 & 32.42 & 3 \\ 
Gemini-1.5-Flash-002~\citep{gemini1.5} &  \underline{40.43}  &  \underline{51.49}  &  \underline{66.00}  &  \underline{54.15}  &  \underline{65.50}  &  \underline{75.49}  & \underline{47.29} & 2 \\ 
Gemini-1.5-Pro-002~\citep{gemini1.5}   &  \textbf{51.91} &  \textbf{60.85}  &  \textbf{73.45}  &  \textbf{67.69}   &  \textbf{72.05}  &  \textbf{82.18}  & \textbf{59.80} & 1 \\ 
\midrule
\multicolumn{9}{c}{\textit{\textbf{Human Performance}}} \\ 
\midrule
Human Experts   & 81.33 & 85.33 & 90.67 & 82.61 & 85.51 & 91.16 & 81.97 & - \\ 
\bottomrule
\end{tabular}
}
\label{tab:long_exp_clean}
\end{table*}

\medskip
\noindent\textbf{Closed-source models outperform open-source models significantly.}
The results demonstrate that closed-source VideoLLMs consistently outperform their open-source counterparts across all montage durations. 
For instance, the best-performing closed-source model (\ie, Gemini-1.5-Pro-002) achieves 65.83\% Recall@1x, tIoU=0.7 on short object needle grounding (Tab.\ \ref{tab:short_exp_clean}) and 48.25\% Recall@1x, tIoU=0.7 on medium object needle grounding (Tab.\ \ref{tab:mid_exp_clean}), surpassing open-source models by a large margin.
{As shown in Tab.\ \ref{tab:long_exp_clean}, we observe that state-of-the-art open-source models (\ie, Time-R1-3B, Time-R1-7B, Qwen2.5-VL-7B, and Qwen2.5-VL-72B) yield near-zero results on long montages, demonstrating their limitations in {retrieval-style long-context recall and temporal grounding}.
In comparison, the best-performing closed-source model (\ie, Gemini-1.5-Pro-002) achieves 51.91\% and 67.69\% Recall@1x, tIoU=0.7 on long object and scene needle grounding, respectively.}

\begin{figure*}[t]
\centering
\includegraphics[width=0.7\textwidth]{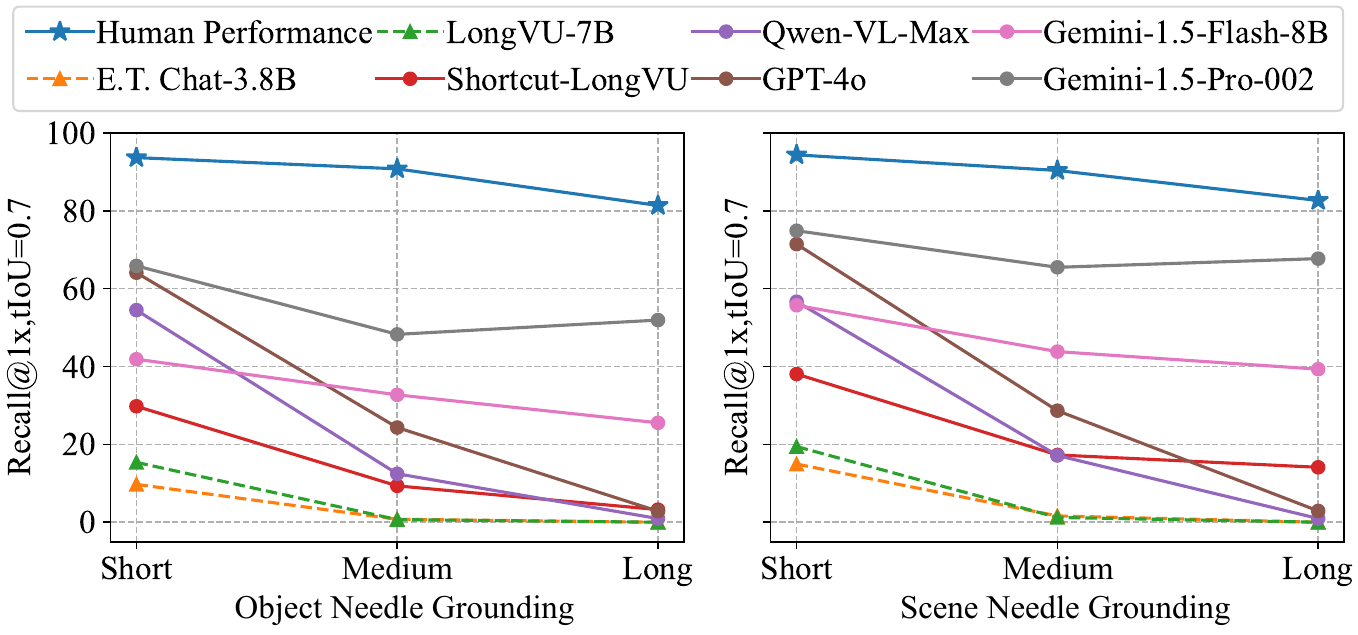}
\caption{Effects of montage duration on \textit{\benchmark{}-Clean}.} 
\label{fig:impact_of_video_duration}
\end{figure*}

\begin{table*}[t!]
\centering
\caption{Evaluation results on single- and multi-needle QA pairs on \textit{\benchmark{}-Clean}.}
\resizebox{0.99\textwidth}{!}{
\begin{tabular}{l|c|cc|cc}
\toprule

\multirow{2}{*}{\textbf{Methods}}  & \multirow{2}{*}{\textbf{\#Needles}} & \multicolumn{2}{c|}{\textbf{Object Needle Grounding}} & \multicolumn{2}{c}{\textbf{Scene Needle Grounding}} \\
\cmidrule{3-6}
& & \textbf{Recall@1x, tIoU=0.7}  & \textbf{Average mAP}  & \textbf{Recall@1x, tIoU=0.7}  & \textbf{Average mAP} \\
\midrule
\multicolumn{6}{c}{\textit{\textbf{Short Montages}}} \\ 
\midrule
\multirow{2}{*}{GPT-4o} & Single & \textbf{73.01} & \textbf{94.48} & 71.29 & \textbf{91.84}  \\
   & Multi & 58.15  &  77.61 & \textbf{71.74} & 86.47   \\ 
\midrule
\multirow{2}{*}{Gemini-1.5-Flash-8B} & Single   & \textbf{62.20} & \textbf{87.40} & \textbf{73.72} & \textbf{92.00}  \\
  & Multi & 27.27 & 46.33 & 41.67 & 60.74  \\ 
\midrule
\multirow{2}{*}{Gemini-1.5-Flash-002} & Single  &  \textbf{68.12} & \textbf{89.49}  & \textbf{74.29}  & \textbf{93.00} \\
   & Multi &  59.01 & 78.87  & 70.47 & 84.05 \\  
\midrule

\multicolumn{6}{c}{\textit{\textbf{Medium Montages}}} \\ \midrule
\multirow{2}{*}{GPT-4o} & Single   &  \textbf{36.48} & \textbf{73.77}  & \textbf{36.00}  & \textbf{76.69}   \\
   & Multi &  17.44  & 47.38  &  23.04  &  54.40   \\ 
\midrule
\multirow{2}{*}{Gemini-1.5-Flash-8B} & Single   & \textbf{47.71}  &  \textbf{71.94}  & \textbf{57.40}  & \textbf{83.08}  \\
  & Multi  & 24.26  & 37.25  &  33.33 & 46.75  \\ 
\midrule
\multirow{2}{*}{Gemini-1.5-Flash-002} & Single  &  \textbf{49.02} & \textbf{76.51}  & \textbf{57.99}  & \textbf{86.43} \\
  & Multi &  37.13 & 57.73  & 53.42 & 71.53 \\  
\midrule

\multicolumn{6}{c}{\textit{\textbf{Long Montages}}} \\ \midrule
\multirow{2}{*}{GPT-4o} & Single   & \textbf{7.14} & \textbf{32.45} & \textbf{7.07} & \textbf{36.77}  \\
   & Multi & 0.00 & 6.10 & 0.00 & 6.90  \\ 
\midrule
\multirow{2}{*}{Gemini-1.5-Flash-8B} & Single   &  \textbf{35.71} &  \textbf{57.04} &  \textbf{56.57} & \textbf{74.85}  \\
& Multi  &  18.25 &  28.81 &  26.15 &  31.09 \\ 
\midrule
\multirow{2}{*}{Gemini-1.5-Flash-002} & Single  & \textbf{46.94}  & \textbf{71.43}  & 51.52  & \textbf{77.78} \\
  & Multi & 35.77  & 55.14  & \textbf{56.15}  & 70.58 \\  
\bottomrule
\end{tabular}
}
\label{tab:single_multi_exp}
\end{table*}

\medskip
\noindent\textbf{Closed-source models still lag substantially behind human experts.}
Despite their overall superior performance compared to open-source counterparts, closed-source models fall significantly short of human expert performance.
Specifically, human experts outperform the best-performing models (\ie, Gemini-1.5-Pro-002) by substantial margins of +27.79\%, +42.50\%, and +29.42\% Recall@1x, tIoU=0.7 on short, medium, and long object needle grounding, respectively.
A similar trend is also observed on scene needle grounding, where human experts consistently surpass both open-source and closed-source models.
The only domain where advanced models approach human-level performance is Average mAP on short object/scene needle grounding, with the gap narrowing to +5.50\%/+3.46\%, reflecting their relatively strong needle detection ability on short montages.
These results highlight the significant challenges faced by models in our \textit{\task{}} task, revealing their limitations in long-text recall and temporal grounding.

\medskip
\noindent\textbf{{Model rankings remain consistent between \textit{\benchmark{}\nobreakdash-Clean} and \textit{\benchmark{}\nobreakdash-Full}.}}
We provide the experimental setups and comprehensive results on the automatically generated \textit{\benchmark{}-Full} in Appendix~\ref{sec:appendix_noisy_benchmark}.
Specifically, experimental results in Tab.~\ref{tab:short_exp_noisy}, Tab.~\ref{tab:mid_exp_noisy}, and Tab.~\ref{tab:long_exp_noisy} show that the model performance rankings of different models remain consistent across these two benchmarks. 
We observe that the rankings exhibit only minor fluctuations of one or two positions among open-source models, while all closed-source models strictly preserve their rankings.
These results highlight that our pipeline can reliably and automatically generate high-quality evaluation data.

\medskip
{
\noindent\textbf{Correlation between \benchmark{} and other benchmarks.}
We analyze the rank correlation between our benchmark and other benchmarks. 
Denote the Pearson correlation coefficient as $r$. 
We present the results as follows: 
$r_\text{E.T. Bench}$ = 0.9591~\cite{etbench},
$r_\text{VideoAutoArena}$ = 0.9055~\cite{VideoAutoArena},
$r_\text{Chatbot Arena-Vision}$ = 0.9516~\cite{arena},
$r_\text{MLVU}$ = 0.9983~\cite{mlvu},
$r_\text{EgoSchema}$ = 0.8619~\cite{egoschema}, 
$r_\text{Video-MME}$ = 0.7771~\cite{videomme}.
All of these results with $p\le 0.05$ indicate that our \textit{\benchmark{}} is strongly correlated with other benchmarks.
These analyses indicate that, although our benchmark is built upon synthetic montages (instead of real-world videos), our \textit{\benchmark{}} can serve as a diagnostic probe that indirectly reflects the video-language understanding capability of advanced VideoLLMs in real-world long videos.
The details of the calculation are provided in the Appendix~\ref{sec:appendix_correlation}.
}

\begin{table*}[t!]
\centering
\caption{Evaluation with different prompts on \textit{\benchmark{}-Clean}.}
\resizebox{\textwidth}{!}{
\begin{tabular}{l|l|ccc}
\toprule
{\textbf{Methods}} & {\textbf{Prompt}} & \textbf{Recall@1x, tIoU0.7} & \textbf{Recall@1x, tIoU0.5} & \textbf{Average mAP} \\
  \midrule
  
\multirow{2}{*}{Qwen-VL-Max} & \textit{video first}   & 28.76   & 39.76   & 69.51       \\
     & \textit{interleaved with timestamp} &  \textbf{54.46} &  \textbf{63.69}      &  \textbf{82.77}       \\

\midrule
\multirow{2}{*}{GPT-4o} & \textit{video first}   & 31.64             & 42.71       & 71.43       \\
       & \textit{interleaved with timestamp} &  \textbf{64.07} &  \textbf{74.85}      &  \textbf{88.56}       \\
\midrule
\multirow{2}{*}{Gemini-1.5-Pro-002}& \textit{video first}            & 35.36    & 48.15      & 76.63       \\
  & \textit{interleaved with timestamp} &  \textbf{65.83} & \textbf{78.22}      &  \textbf{89.81}       \\ 

\bottomrule

\end{tabular}
}
\label{tab:prompts_exp}
\end{table*}

\subsection{Further Analyses}
\label{sec:exp_analysis}

\noindent\textbf{Effects of montage duration.}
As shown in Fig.~\ref{fig:impact_of_video_duration}, the performance (using Recall@1x, tIoU=0.7) of VideoLLMs drops significantly as the montage duration increases.
Open-sourced VideoLLMs are particularly affected by longer montage durations, indicating that their temporal grounding capabilities on longer videos require further improvement.
Similarly, Qwen-VL-Max and GPT-4o struggle with longer videos due to their limited context lengths, resulting in the neglect of valuable contextual video information.
The performance of Gemini-1.5-Flash-8B also drops significantly as the montage duration increases, which may be attributed to its relatively smaller model size.
An exception to this trend is Gemini-1.5-Pro-002, which achieves comparable performance on both medium and long montages, reflecting its robustness in handling long visual sequences.

\medskip
\noindent\textbf{Effects of the number of target needles.}
Tab.~\ref{tab:single_multi_exp} shows the experimental results on short montages, where VideoLLMs consistently underperform on multi-needle grounding compared with single-needle grounding.
These findings reveal the substantial challenges posed by our multi-needle setting, illustrating the varying levels of QA difficulty on \textit{\benchmark{}}.

\medskip
\noindent\textbf{Effects of prompt choice.}
Tab.~\ref{tab:prompts_exp} presents the results on short object needle grounding, indicating that models utilizing the \textit{interleaved with timestamp} prompt significantly outperform those using the \textit{video first} prompt.
A possible reason is that the former provides a clearer frame-timestamp correspondence, which plays a vital role in assisting models to effectively comprehend and ground video content.

\medskip
\noindent\textbf{Effects of models' input frame capacity.}
As shown in Tab.~\ref{tab:short_exp_clean}, GPT-4o outperforms Qwen-VL-Max on short montages, despite both models having the same input frame capacity of 1 FPS.
Tab.~\ref{tab:mid_exp_clean} and Tab.~\ref{tab:long_exp_clean} further reveal that, although Qwen-VL-Max has more than twice the input frame capacity of GPT-4o on longer montages, its performance still lags behind that of GPT-4o. 
These observations suggest that merely increasing the input frame capacity does not inherently guarantee improved temporal grounding ability.

\medskip
\noindent\textbf{Effects of LLM model size.}
Tab.~\ref{tab:effect_llm_size} presents the effect of different LLM sizes (using the same vision encoder) on short object needle grounding.
The results demonstrate that VideoLLMs with larger LLMs significantly outperform those with smaller ones, 
indicating that a more capable LLM enhances VideoLLMs' temporal {understanding} capability.
This observation further suggests that the superior performance of closed-source models on \textit{\benchmark{}} may be attributed to their more powerful LLM backbones.

\medskip
\noindent\textbf{Discussion on shortcut baseline.}
To better analyze the performance of open-source VideoLLMs, we design a shortcut baseline tailored to our task.
Rather than processing an entire long-context montage at once, the shortcut baseline (based on LongVU-7B) first segments the input into shorter video clips and then performs temporal grounding sequentially.
Experimental results show that the shortcut baseline outperforms open-source VideoLLMs across different montage durations (\eg, achieving +14.40\%/+8.66\% Recall@1x, tIoU=0.7 on short/medium object needle grounding compared with LongVU-7B).
However, the shortcut baseline still significantly lags behind the closed-source VideoLLMs (\eg, achieving -36.06\%/-38.90\% Recall@1x, tIoU=0.7 on short/medium object needle grounding compared with Gemini-1.5-Pro-002), underscoring the need to further enhance the inherent temporal grounding capability of open-source VideoLLMs.

\begin{table*}[t!]
\centering
\caption{Effects of varying LLM model sizes with the same vision encoder on \textit{\benchmark{}-Clean}. The LLaVA-Video model series employ SigLIP-400M~\citep{siglip} as the vision encoder, while the Time-R1 and Qwen2.5-VL model series adopt a customized ViT~\citep{qwen2-vl}.}
\resizebox{\textwidth}{!}{
\begin{tabular}{l|l|ccc}
\toprule
{\textbf{Methods}} & {\textbf{LLM}} & \textbf{Recall@1x, tIoU0.7} & \textbf{Recall@1x, tIoU0.5} & \textbf{Average mAP} \\
  \midrule
  
LLaVA-Video-7B & Qwen2-7B~\citep{DBLP:journals/corr/abs-2407-10671qwen2}   & 1.10   & 1.90 & 11.27       \\
LLaVA-Video-72B & Qwen2-72B~\citep{DBLP:journals/corr/abs-2407-10671qwen2} &  \textbf{8.88} (+7.78) &  \textbf{18.36} (+16.46) &  \textbf{34.95}  (+23.68)     \\

\midrule

TimeR1-3B & Qwen2.5-7B~\citep{DBLP:journals/corr/abs-2412-15115qwen25}   & 15.27 & 32.53 & 63.84       \\
TimeR1-7B & Qwen2.5-72B~\citep{DBLP:journals/corr/abs-2412-15115qwen25} & \textbf{22.36} (+7.09) &  \textbf{38.02} (+5.49)     &  \textbf{68.25} (+4.41)      \\

\midrule

Qwen2.5-VL-7B & Qwen2.5-7B~\citep{DBLP:journals/corr/abs-2412-15115qwen25}   & 24.45 & 32.63       & 57.22       \\
Qwen2.5-VL-72B & Qwen2.5-72B~\citep{DBLP:journals/corr/abs-2412-15115qwen25} & \textbf{28.04} (+3.59)  &  \textbf{37.62} (+4.99)      &  \textbf{61.01} (+3.79)      \\

\bottomrule

\end{tabular}
}
\label{tab:effect_llm_size}
\end{table*}

\begin{table*}[ht]
\centering
\caption{{Model performance across various needle positions on \textit{\benchmark{}-Clean} (long montages). 
\textbf{Avg}: Average Recall@1x, tIoU=0.7 across object and scene needle grounding. \textbf{Bold}: Best non-zero result across beginning range (0--33\%), middle range (33--67\%), and end range (67--100\%).}}
\renewcommand\arraystretch{0.92}
\resizebox{\textwidth}{!}{
\setlength{\tabcolsep}{1.5mm}{
\begin{tabular}{l|c|ccc|ccc|c}
\toprule
\multirow{3}{*}{\textbf{Methods}} & \multirow{3}{*}{\textbf{Position}} & \multicolumn{3}{c|}{\textbf{Long Object Needle Grounding}} & \multicolumn{3}{c|}{\textbf{Long Scene Needle Grounding}} & \multirow{3}{*}{\textbf{Avg}} \\
\cmidrule{3-8}
& & \textbf{Recall@1x,} & \textbf{Recall@1x,} & \textbf{Average} & \textbf{Recall@1x,} & \textbf{Recall@1x,} & \textbf{Average} \\
& & \textbf{tIoU=0.7} & \textbf{tIoU=0.5} & \textbf{mAP} & \textbf{tIoU=0.7} & \textbf{tIoU=0.5} & \textbf{mAP} \\
\midrule
\multicolumn{9}{c}{\textit{\textbf{Open-Source VideoLLMs}}} \\ 
\midrule
\multirow{3}{*}{Time-R1-3B} & 0--33\% & 0.00 & 0.00 & 0.29 & 0.00 & 0.00 & 0.37 & 0.00 \\
 & 33--67\% & 0.00 & 0.00 & 0.00 & 0.00 & 0.00 & 0.00 & 0.00 \\
 & 67--100\% & 0.00 & 0.00 & 0.36 & 0.00 & 0.00 & 0.29 & 0.00 \\
\midrule
\multirow{3}{*}{Time-R1-7B} & 0--33\% & 0.97 & 2.91 & 6.12 & 0.00 & 3.00 & 5.90 & 0.49 \\
 & 33--67\% & \textbf{1.64} & 1.64 & 2.95 & 0.00 & 0.00 & 0.00 & \textbf{0.82} \\
 & 67--100\% & 0.00 & 0.00 & 0.36 & 0.00 & 0.00 & 1.36 & 0.00 \\

\midrule
\multirow{3}{*}{Qwen2.5-VL-7B} & 0--33\% & \textbf{0.97} & 1.94 & 3.09 & \textbf{1.00} & 2.00 & 2.14 & \textbf{0.99} \\
 & 33--67\% & 0.00 & 0.00 & 0.40 & 0.00 & 0.00 & 0.24 & 0.00 \\
 & 67--100\% & 0.00 & 0.00 & 1.04 & 0.00 & 0.00 & 0.85 & 0.00 \\

\midrule
\multirow{3}{*}{Qwen2.5-VL-72B} & 0--33\% & \textbf{0.97} & 1.94 & 2.32 & \textbf{1.00} & 3.00 & 4.08 & \textbf{0.99} \\
 & 33--67\% & 0.00 & 0.00 & 0.00 & 0.00 & 0.00 & 0.00 & 0.00 \\
 & 67--100\% & 0.00 & 0.00 & 0.80 & 0.00 & 0.00 & 0.76 & 0.00 \\

\midrule
\multicolumn{9}{c}{\textit{\textbf{Closed-Source VideoLLMs}}} \\ 

\midrule
\multirow{3}{*}{Qwen-VL-Max} & 0--33\% & 1.09 & 3.26 & 6.33 & 1.11 & 5.56 & 9.22 & 1.10 \\
 & 33--67\% & 0.00 & 0.00 & 2.52 & 0.00 & 1.96 & 2.70 & 0.00 \\
 & 67--100\% & \textbf{1.45} & 1.45 & 1.44 & \textbf{1.43} & 1.43 & 1.80 & \textbf{1.44} \\
\midrule
\multirow{3}{*}{GPT-4o} & 0--33\% & 2.88 & 5.77 & 15.77 & 3.00 & 7.00 & 19.93 & 2.94 \\
 & 33--67\% & 1.61 & 9.68 & 20.70 & 1.67 & 8.33 & 20.17 & 1.64 \\
 & 67--100\% & \textbf{3.70} & 7.41 & 14.16 & \textbf{3.70} & 8.64 & 17.37 & \textbf{3.70} \\

\midrule
\multirow{3}{*}{Gemini-1.5-Flash-8B} & 0--33\%  & 20.00          & 32.00 & 30.13 & 26.53          & 30.61 & 31.19 & 23.27 \\
 & 33--67\% & \textbf{21.05} & 26.32 & 26.54 & \textbf{28.13} & 28.13 & 29.95 & \textbf{24.59} \\
 & 67--100\% & 14.29          & 16.33 & 26.43 & 24.49          & 26.53 & 29.86 & 19.39 \\

\midrule
\multirow{3}{*}{Gemini-1.5-Flash-002} & 0--33\% & 43.43          & 55.56 & 63.96 & \textbf{58.33} & 64.58 & 74.79 & \textbf{50.88} \\
 & 33--67\% & \textbf{45.76} & 55.93 & 65.06 & 52.63          & 66.67 & 75.18 & 49.20 \\
 & 67--100\% & 32.47          & 42.86 & 55.54 & 50.00          & 65.79 & 71.25 & 41.23 \\

\midrule
\multirow{3}{*}{gemini-1.5-pro-002} & 0--33\% & \textbf{53.54} & 63.64 & 69.23 & 63.54          & 66.67 & 82.36 & 58.54 \\
 & 33--67\% &  49.15          & 61.02 & 72.82 & \textbf{73.68} & 80.70 & 84.80 & \textbf{61.42} \\
 & 67--100\% & 51.95          & 57.14 & 73.64 & 68.42          & 72.37 & 80.79 & 60.18 \\
\bottomrule
\end{tabular}
}
}
\label{tab:needle_position_bias_long}
\end{table*}

{
\medskip
\noindent\textbf{Effects of Needle Positions.}
We sincerely thank the reviewer for the constructive suggestion regarding the sensitivity of model performance to different needle positions within the montage.
Tab.~\ref{tab:needle_position_bias_long} presents the sensitivity of model performance to different needle positions within the montage.
Specifically, following the setting of MVBench~\citep{mvbench}, we divide the needle positions into three temporal ranges according to their relative locations within the montage: beginning (0--33\%), middle (33--67\%), and end (67--100\%). 

The experimental results show that the model performance (using the Avg metric) on our benchmark is indeed sensitive to the needle positions within the montage.
However, we would like to clarify that the observed positional bias is model-dependent, and does not indicate a consistent preference for either the beginning or the end across all models.
For example, Qwen2.5-VL-7B, Qwen2.5-VL-72B, and Gemini-1.5-Flash-002 achieve the best performance when the needle appears in the beginning range.
In contrast, Time-R1-7B, Gemini-1.5-Flash-8B, and Gemini-1.5-Pro-002 perform best when the needle appears in the middle range.
Meanwhile, Qwen-VL-Max and GPT-4o perform best when the needle appears near the end of the montage.
These results suggest that there is no uniform beginning- or end-position bias across both open-source and closed-source models.

We also observe that the positional bias can vary across different needle types (using the Recall@1x at tIoU=0.7 metric).
For instance, Gemini-1.5-Flash-002 achieves its best object needle grounding performance in the middle range, while achieving its best scene needle grounding performance in the beginning range.
Similarly, Gemini-1.5-Pro-002 performs best on object needle grounding in the beginning range, while its best scene needle grounding performance appears in the middle range.
These results suggest that positional sensitivity is not only model-dependent but also affected by the type of needle to be retrieved and temporally grounded.

Moreover, we observe that stronger models tend to be more robust (\ie, less sensitive) to different needle positions.
For example, the best model (\ie, Gemini-1.5-Pro-002) exhibits the most stable performance across the beginning, middle, and end ranges, with the Avg metrics of 58.54\%, 61.42\%, and 60.18\%, respectively.
In comparison, less robust models exhibit larger positional fluctuations.
For instance, Gemini-1.5-Flash-002, Gemini-1.5-Flash-8B, and GPT-4o exhibit a clear performance gap between one temporal range and the other two ranges, while Qwen-VL-Max and all open-source models even achieve zero performance in certain temporal ranges.

Overall, these results indicate that both open-source and closed-source models exhibit model-dependent positional biases, reflecting their varying capabilities in retrieval-style long-context recall and temporal grounding.
Moreover, we also provide additional experiments on needle relocation in Appendix~\ref{sec:appendix_needle_relocation}.

}

\begin{figure*}[t]
\centering
\includegraphics[width=\textwidth]{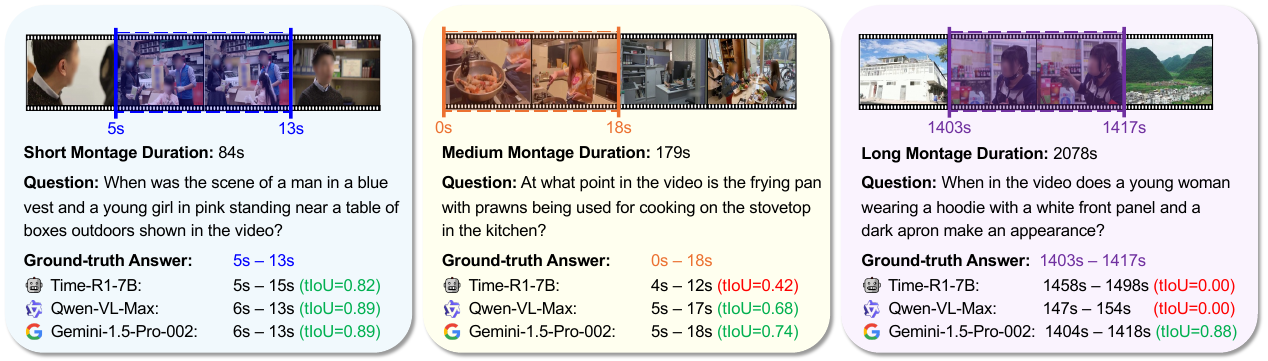}
\caption{Qualitative comparisons of needle grounding questions across varying montage durations from \textit{\benchmark{}-Clean} (see the details of tIoU in Appendix~\ref{sec:appendix_evaluation_metric}).
}
\label{fig:case_duration}
\end{figure*}

\begin{figure*}[t]
\centering
\includegraphics[width=\textwidth]{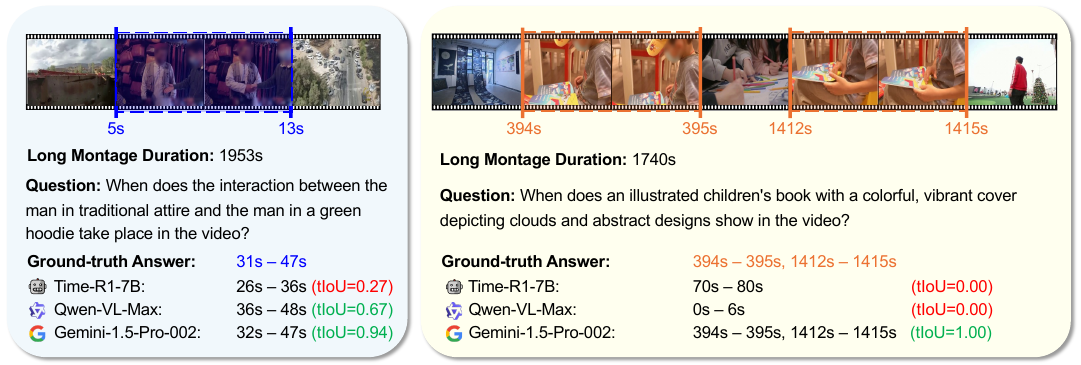}
\caption{Qualitative comparisons of needle grounding questions with single target needle (left) and multiple target needles (right) from \textit{\benchmark{}-Clean} (long montages).}
\label{fig:case_num_needle}
\end{figure*}

\subsection{Visualization and Case Study}
\label{sec:visualization}
In this section, we present qualitative comparison results to demonstrate several dimensions that affect the QA difficulty on our benchmark. 
Specifically, for each example, we visualize the montage, the needle grounding QA pair, and the corresponding predictions of three state-of-the-art models, along with their tIoU scores.

\begin{figure*}[t]
\centering
\includegraphics[width=\textwidth]{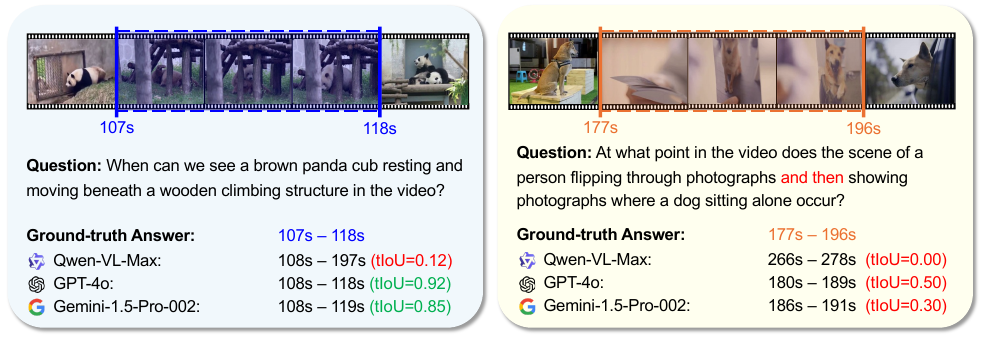}
\caption{Qualitative comparisons of needle grounding questions without temporal qualifiers (left) and with temporal qualifiers (right) from \textit{\benchmark{}-Clean} (medium montages), reflecting the different levels of difficulty in temporal understanding.} 
\label{fig:case_question}
\end{figure*}

\medskip
\noindent\textbf{Montage duration.}
Fig.~\ref{fig:case_duration} presents that montage duration exerts a substantial influence on model performance.
In the short-montage example, all three models consistently achieve high tIoU scores (\ie, tIoU$>$0.7).
However, as the montage duration increases, our \textit{\task{}} task becomes more challenging.
Both Time-R1-7B and Qwen-VL-Max show consistent degradation on longer montages, where Time-R1-7B reaches a low tIoU (\ie, tIoU$<$0.5) in the medium-montage example, and both models even yield zero tIoU in the long-montage example.
By contrast, Gemini-1.5-Pro-002 achieves high tIoU in both medium- and long-montage examples, due to its stronger robustness across different montage durations.

\medskip
\noindent\textbf{Number of target needles.}
As shown in Fig.~\ref{fig:case_num_needle}, the results indicate that a multi-needle question demands more sophisticated temporal grounding than a single-needle question in the proposed \textit{\task{}} task.
Specifically, we observe that both Time-R1-7B and Qwen-VL-Max are limited to single-needle predictions and exhibit significant performance drops in the multi-needle example, with tIoU scores even dropping to zero.
In comparison, Gemini-1.5-Pro-002 accurately handles the multi-needle grounding question, reflecting its stronger temporal grounding capability.

\medskip
\noindent\textbf{Temporal qualifiers.}
In Fig.~\ref{fig:case_question}, we observe that needle grounding questions involving temporal qualifiers (\eg, ``then'', ``before'', ``after'') demand {more advanced temporal understanding (\ie, a certain level of temporal reasoning)} in the proposed \textit{\task{}} task.
In the example without temporal qualifiers, all three models roughly cover the ground-truth interval in their predictions, with GPT-4o and Gemini-1.5-Pro-002 achieving high tIoU, while Qwen-VL-Max predicts an over-extended span and attains a low tIoU.
In contrast, these models exhibit consistent performance drops in the example with temporal qualifiers (\ie, ``and then''). 
To be specific, GPT-4o and Gemini-1.5-Pro-002 capture only a portion of the ground-truth interval and thus yield relatively low tIoU, while Qwen-VL-Max fails to capture the target needle (\ie, tIoU=0.0).
These observations reveal the limitations of advanced models in temporal understanding, highlighting a clear direction for future model development.

\section{Conclusion and Future Work}
In this paper, we introduce an innovative video-language task, {``needle in a montage''}.
Our task is tailored for {evaluating advanced VideoLLMs' temporal understanding abilities, including retrieval-style long-context recall and temporal grounding}.
We further propose an automated data generation pipeline, facilitating high-quality data synthesis.
Leveraging the proposed pipeline, we establish \textit{\benchmark{}}, a benchmark centered on {``needle in a montage''}.
Extensive experiments conducted on our benchmark demonstrate the significant challenges posed by our task, 
while providing valuable insights into the capacities and limitations of existing VideoLLMs.

\smallskip
\noindent{\textbf{Future work.}}
Moving forward, we plan to continue expanding the benchmark and updating the leaderboard to evaluate emerging VideoLLMs. Another promising avenue we will explore is to further refine the proposed automated data generation pipeline and generalize it into a versatile framework for multimodal evaluation. By reducing manual annotation costs and enabling scalable, high-quality test data creation, this direction has the potential to transform how multimodal systems are evaluated across domains.
{Given the scalability and cost-effectiveness of our automated pipeline, it has the potential to create large-scale training data to enhance models' long-context and temporal grounding capabilities.}

\smallskip
\noindent{\textbf{Limitations.}}
{
Despite our best efforts, we acknowledge that there are still some limitations in our work.
Firstly, our proposed automated pipeline relies on a single annotation model (\ie, GPT-4o), which may introduce potential generation bias and benchmarking favor. 
To mitigate these issues, our pipeline can be extended to incorporate multiple annotation models.
Secondly, the authorized videos used in our benchmark are curated from a single media platform (\ie, Phoenix TV), which may introduce a degree of domain bias, particularly toward professionally produced broadcast content with consistent cinematography and visual style.
Continuously incorporating more diverse videos from multiple authorized platforms represents an important direction for the future development and generalization of our benchmark.
Thirdly, our \textit{\task{}} task is built upon synthetic montages and thus does not fully capture the specific challenges posed by real-world long videos, such as temporal reasoning and temporal-coherence-based video understanding.
}

\section*{Declarations}
\noindent\textbf{Data availability} Our benchmark and model evaluation service will be made publicly available on requests for research purposes.

\smallskip
\noindent\textbf{Conflict of interest} The authors declare that they have no conflict of interest.

\smallskip
\noindent\textbf{Consent for publication}  All authors consent to see their work published if accepted.

\bibliography{sn-bibliography}

\newpage
\appendix

\section*{Appendix}

\section{Benchmark Statistics}
\label{sec:appendix_benchmark_statistics}
\begin{figure*}[htbp]
  \centering
  \begin{subfigure}[b]{0.33\textwidth}
    \includegraphics[width=\linewidth]{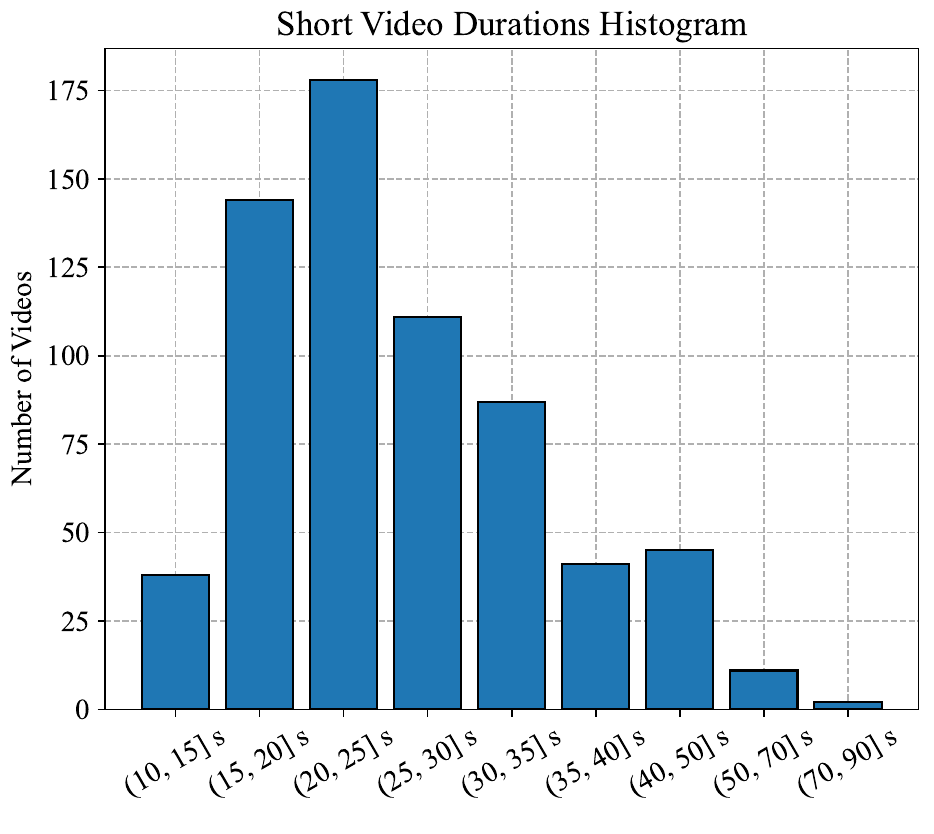}
    \caption{Short Montage Durations Histogram}
  \end{subfigure}
  \begin{subfigure}[b]{0.33\textwidth}
    \includegraphics[width=\linewidth]{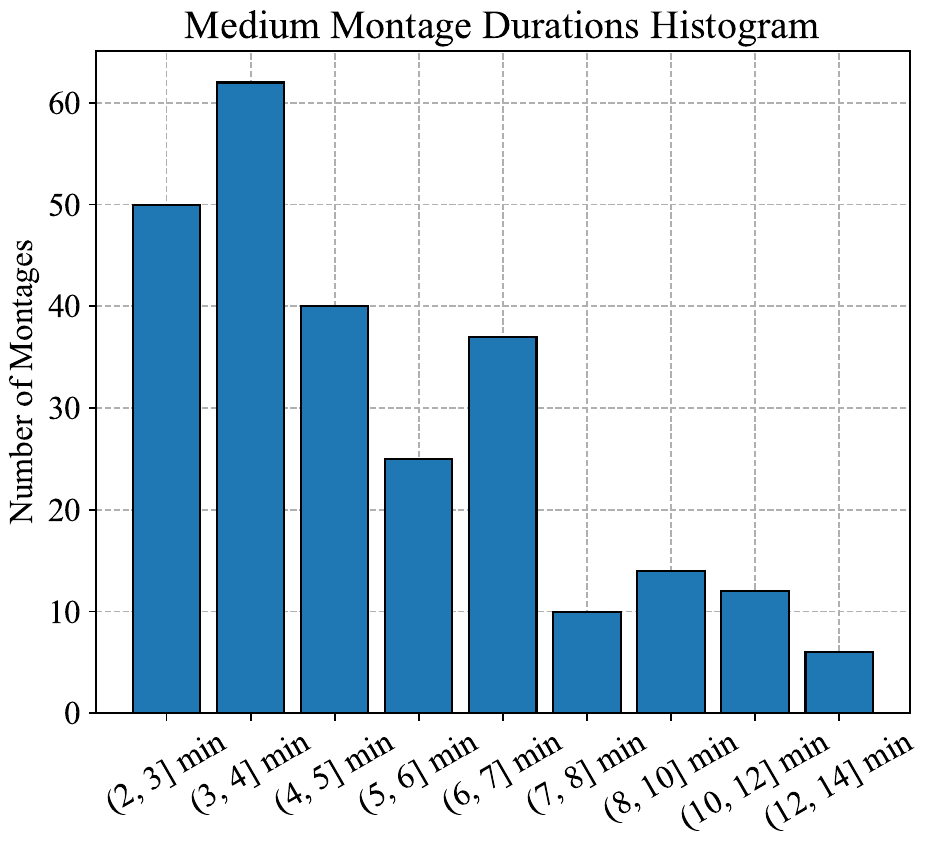}
    \caption{Medium Montage Durations Histogram}
  \end{subfigure}
  \begin{subfigure}[b]{0.32\textwidth}
    \includegraphics[width=\linewidth]{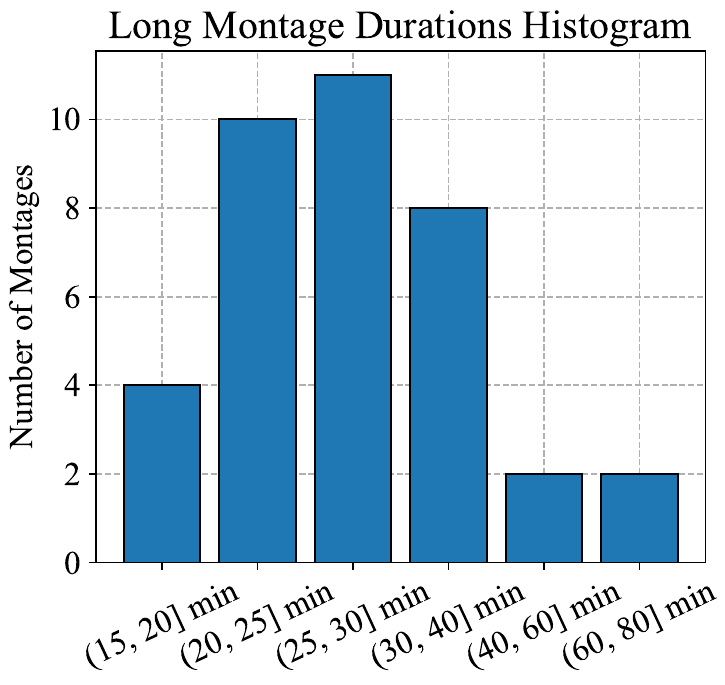}
    \caption{Long Montage Durations Histogram}
  \end{subfigure}
  
  \caption{The distribution of montage duration \textit{\benchmark{}-Clean}.} 
  \label{fig:video_durations}
\end{figure*}

\begin{figure}[ht]
\centering
\includegraphics[width=\columnwidth]{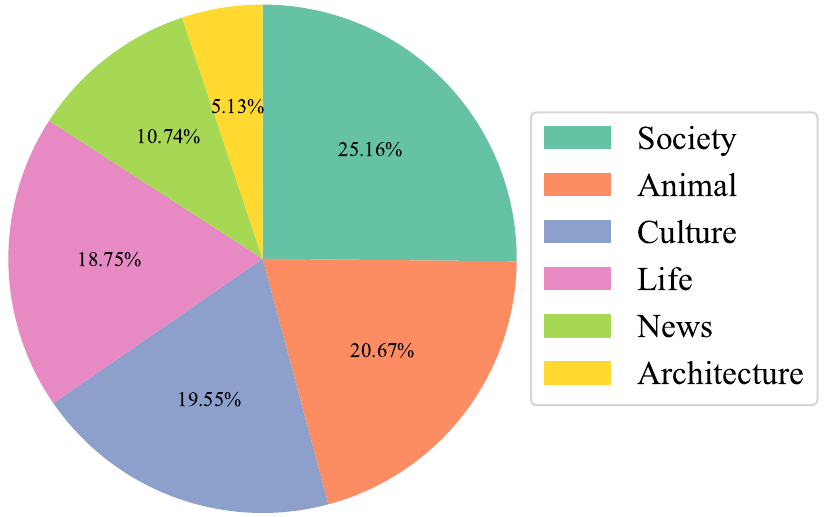}
  \caption{The distribution of video categories on \textit{\benchmark{}-Clean}.} 
\label{fig:video_category}
\end{figure}

\begin{figure*}[p]
\centering
\includegraphics[height=0.54\textwidth, angle=90]{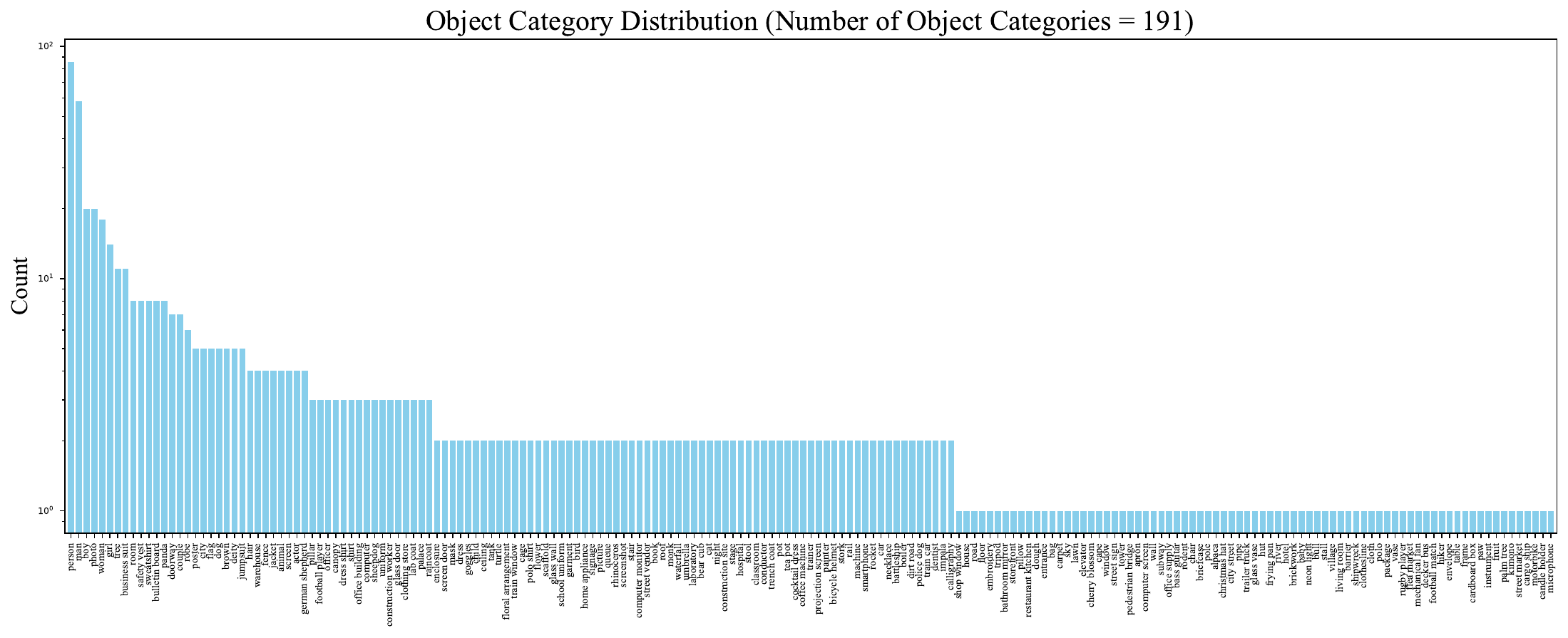}
  \caption{The distribution of target object categories within \textit{\benchmark{}-Clean}.} 
\label{fig:object_category}
\end{figure*}

Tab.~\ref{tab:benchmark_full_clean} presents the statistics of different benchmark variants, including \textit{\benchmark{}-Full} and \textit{\benchmark{}}.
Additionally, Fig.~\ref{fig:video_durations}, Fig.~\ref{fig:video_category}, and Fig.~\ref{fig:object_category} present the distributions of montage durations, video categories, and target object categories within our \textit{\benchmark{}}, respectively.

\section{Details of Data Generation Pipeline Analysis}
\label{sec:appendix_gen_data_compare}
\noindent\textbf{Conventional Manual Annotation Process (\ie, ManualAnno).}
Given a video with $T$ seconds, the conventional annotation process from scratch (refer to conventional temporal grounding works~\citep{lei2021qv}) goes through three steps: 

(1) \textit{Question construction phase}: A human annotator watches the video for the first time and constructs a meaningful natural language temporal grounding question describing the target object/scene (referring to single or multiple target video clips within the video). 
This phase requires $T+S_1$ seconds, where $S_1$ includes reading the annotation instruction, recalling the target video clips where the target object/scene appears, and constructing the corresponding natural language question.

(2) \textit{Answer construction phase}: The same human annotator watches the video for the second time to find and annotate the ground-truth answer of the natural language question (\ie, the temporal extents of all target video clips within the video). 
This phase will take $T+S_2$ seconds, where $S_2$ includes reading the annotation instruction and finding and annotating the ground-truth answer.

(3) \textit{Cleaning phase}: Another human annotator watches the unseen video, and refines both the annotated question and the annotated answer. 
This phase will take $T+S_3$ seconds, where $S_3$ includes reading the annotation instruction and refining the annotated question and the annotated answer. 

To summarize, the manual annotation time cost of \textit{ManualAnno} can be formulated as:
\begin{equation}
    T_{\textit{ManualAnno}} = 3T+S_1+S_2+S_3.
    \label{eq:t_human}
\end{equation}
In practice, $S_2$ and $S_3$ scale with the number of target video clips.

\medskip
\noindent\textbf{Our Automated Data Generation Pipeline (\ie, {NeMo-Auto}).}
The raw QA pairs of our \textit{\benchmark{}-Clean} are automatically generated using the proposed data generation pipeline, subsequently involving a minor amount of manual effort for QA verification to ensure a precise, robust model benchmarking.
Therefore, the only manual step in \textit{{NeMo-Auto}} is the manual QA verification (\ie, \textit{Cleaning phase}).
Similarly, this step will take $T+S_3'$ seconds, where $S_3'$ includes reading the annotation instruction and refining the annotated question and the annotated answer. 

In summary, the manual annotation time cost of \textit{{NeMo-Auto}} can be formulated as:
\begin{equation}
    T_{\textit{NeMo-Auto}} = T+S_3'.
    \label{eq:t_auto}
\end{equation}

\medskip
\noindent\textbf{Theoretical Cost Reduction.}
The time reduction ratio between \textit{ManualAnno} and \textit{{NeMo-Auto}} can be formulated as:
\begin{equation}
    \alpha = 1 - \frac{T_{\textit{NeMo-Auto}}}{T_{\textit{ManualAnno}}} = \frac{2T+S_1+S_2+S_3-S_3'}{3T+S_1+S_2+S_3}.
\end{equation}
Generally, QA annotation demands more human cognitive effort than QA cleaning (\ie, $S_1 \gg S_3$, $S_2 > S_3$). 
For short- and medium-duration videos, QA annotation typically consumes more time than video watching (\ie, $S_1 > T$, $S_2 > T$). 
Conversely, for long-duration videos, video watching requires more time than QA annotation (\ie, $T \gg S_1 + S_2$).
Compared to Eq.~\ref{eq:t_human} and Eq.~\ref{eq:t_auto}, \textit{ManualAnno} requires significantly more time for video watching than our data generation pipeline (\ie, $3T > T$), while both methods exhibit similar time cost for QA cleaning (\ie, $S_3 \approx S_3'$). 

Considering these observations, the manual time reduction ratio for different video durations can be approximated as follows:
\begin{equation}
    \alpha \approx \frac{2T+S_1+S_2}{3T+S_1+S_2} = \frac{2+\frac{S_1+S_2}{T}}{3+\frac{S_1+S_2}{T}},
\end{equation}
where $T > 0$, $S_1 > 0$, and $S_2 > 0$, resulting in $\alpha(T, S_1, S_2)$ being a monotonically increasing function with respect to these variables. For short- and medium-duration videos, where $\frac{S_1+S_2}{T} > 2$, we have $\alpha_\text{short} \approx \alpha_\text{medium} > 0.8$. 
For long-duration videos, where $T \gg (S_1+S_2)$, we have $\alpha_\text{long} \approx 0.67$. 

Therefore, the average time efficiency of \textit{{NeMo-Auto}} with respect to \textit{ManualAnno} is computed as:
\begin{equation}
\alpha_\text{average} = (\alpha_\text{short}+\alpha_\text{medium}+\alpha_\text{long})/3 > 0.76.
\end{equation}

\begin{table*}[t!]
\centering
\caption{The statistics of \textit{\benchmark{}}.}
\resizebox{\textwidth}{!}
{
\begin{tabular}{l|c|c|c|c}
\toprule
\multirow{2}{*}{\textbf{Benchmark}} & \textbf{Short} & \textbf{Medium} & \textbf{Long} & \textbf{All}\\
& \textbf{\#QA Pairs / \#Montage} & \textbf{\#QA Pairs / \#Montage} & \textbf{\#QA Pairs / \#Montage} & \textbf{\#QA Pairs / \#Montage} \\

\midrule
\textit{\benchmark{}-Full}  & 12,932 / 6,466 & 6,760 / 3,380 & 11,686 / 3,640 & 31,378 / 13,486 \\    
\midrule
\textit{\benchmark{}-Clean} & 1,246 / 657 & 507 / 256 & 300 / 37 & 2,053 / 940\\
\bottomrule
\end{tabular}
}
\label{tab:benchmark_full_clean}
\end{table*}

\section{Details of Experiments}
\label{sec:appendix_setup}

\subsection{Details of Prompt Design}
\label{sec:appendix_prompt_design}
\begin{figure}[t!]
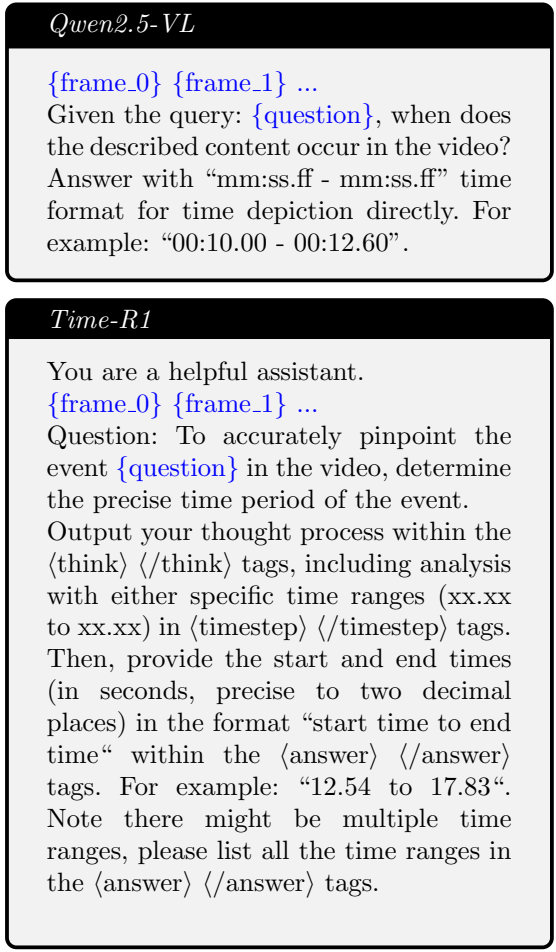

\begin{center}
\begin{tcolorbox}[colback=black!5!white, colframe=black!75!black, title=\textit{Qwen2.5-VL}, width=0.95\linewidth]
\textcolor{blue}{\{frame\_0\} \{frame\_1\} ...} \\
Given the query: \textcolor{blue}{\{question\}}, when does the described content occur in the video? Answer with ``mm:ss.ff - mm:ss.ff'' time format for time depiction directly. For example: ``00:10.00 - 00:12.60''.
\end{tcolorbox}
\begin{tcolorbox}[colback=black!5!white, colframe=black!75!black, title=\textit{Time-R1}, width=0.95\linewidth]
You are a helpful assistant.\\
\textcolor{blue}{\{frame\_0\} \{frame\_1\} ...} \\
Question: To accurately pinpoint the event \textcolor{blue}{\{question\}} in the video, determine the precise time period of the event. \\
Output your thought process within the $\langle$think$\rangle$ $\langle$/think$\rangle$ tags, including analysis with either specific time ranges (xx.xx to xx.xx) in $\langle$timestep$\rangle$ $\langle$/timestep$\rangle$ tags. \\
Then, provide the start and end times (in seconds, precise to two decimal places) in the format ``start time to end time`` within the $\langle$answer$\rangle$ $\langle$/answer$\rangle$ tags. For example: ``12.54 to 17.83``. Note there might be multiple time ranges, please list all the time ranges in the $\langle$answer$\rangle$ $\langle$/answer$\rangle$ tags. \\
\end{tcolorbox}
\caption{Prompts for Qwen2.5-VL and Time-R1 on \textit{\benchmark{}}, which are modified from their official implementations.}
\label{fig:other_prompt_design}
\end{center}
\end{figure}

\begin{figure}[t!]
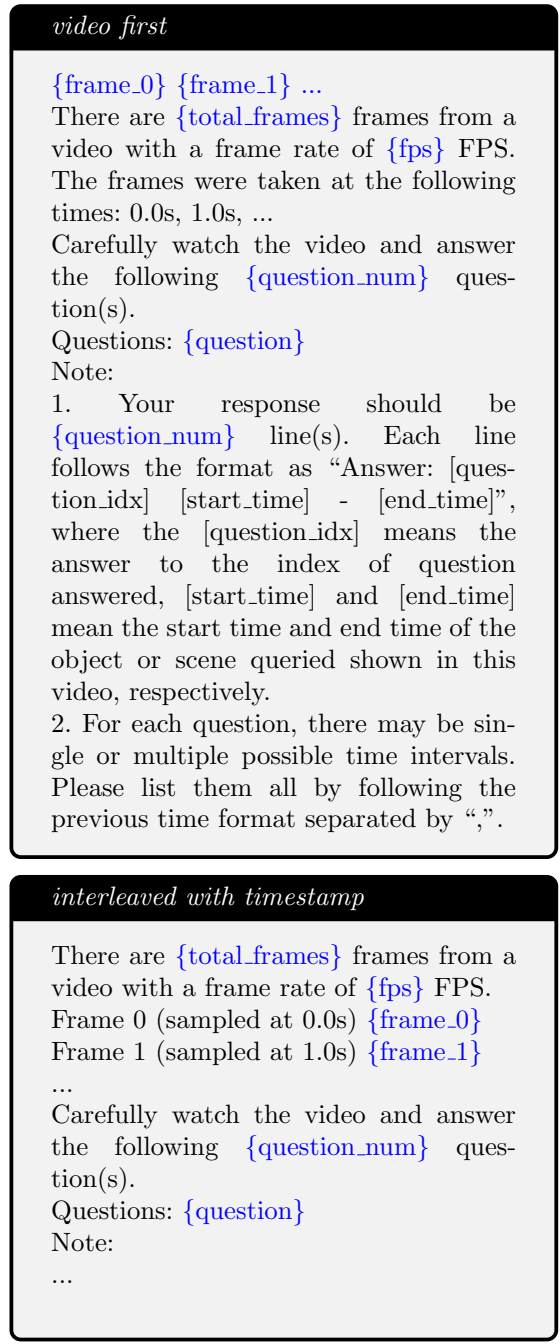

\begin{center}
\begin{tcolorbox}[colback=black!5!white, colframe=black!75!black, title=\textit{video first}, width=0.95\linewidth]
\textcolor{blue}{\{frame\_0\} \{frame\_1\} ...} \\
There are \textcolor{blue}{\{total\_frames\}} frames from a video with a frame rate of \textcolor{blue}{\{fps\}} FPS. 
The frames were taken at the following times: 0.0s, 1.0s, ... \\
Carefully watch the video and answer the following \textcolor{blue}{\{question\_num\}} question(s).\\
Questions: \textcolor{blue}{\{question\}} \\
Note: \\
1. Your response should be \textcolor{blue}{\{question\_num\}} line(s). 
Each line follows the format as ``Answer: {[question\_idx]} {[start\_time]} - {[end\_time]}'',
where the {[question\_idx]} means the answer to the index of question answered, {[start\_time]} and {[end\_time]} mean the start time and end time of the object or scene queried shown in this video, respectively. \\
2. For each question, there may be single or multiple possible time intervals. Please list them all by following the previous time format separated by ``,''. 
\end{tcolorbox}

\begin{tcolorbox}[colback=black!5!white, colframe=black!75!black, title=\textit{interleaved with timestamp}, width=0.95\linewidth]
There are \textcolor{blue}{\{total\_frames\}} frames from a video with a frame rate of \textcolor{blue}{\{fps\}} FPS. \\
Frame 0 (sampled at 0.0s) \textcolor{blue}{\{frame\_0\}} \\
Frame 1 (sampled at 1.0s) \textcolor{blue}{\{frame\_1\}} \\
... \\
Carefully watch the video and answer the following \textcolor{blue}{\{question\_num\}} question(s).\\
Questions: \textcolor{blue}{\{question\}} \\
Note: \\
... \\
\end{tcolorbox}

\caption{Batch inference prompts for closed-source VideoLLMs.}
\label{fig:batch_prompts}
\end{center}
\end{figure}

For Qwen2.5-VL-7B, Qwen2.5-VL-72B, Time-R1-3B, and Time-R1-7B, we adopt the temporal grounding prompts from their official implementations, as shown in Fig.~\ref{fig:other_prompt_design}.
Motivated by~\citep{chandrasegaran2024hourvideo}, we employ a batch inference strategy for closed-source VideoLLMs on medium- and long-duration montages to reduce evaluation API costs. The batch inference prompts are illustrated in Fig.~\ref{fig:batch_prompts}.
In addition, we have investigated the ``Socratic models'' approach following~\citep{chandrasegaran2024hourvideo,DBLP:conf/iclr/ZengAICWWTPRSLV23}, where minute-level video captions are provided as additional model input.
However, our preliminary experiments suggest that the ``Socratic models'' approach does not yield better results.
Therefore, we do not consider it in this work.

\subsection{Evaluation Protocol.}
To ensure precise assessment of VideoLLMs, all experiments are conducted on the rigorously verified \textit{\benchmark{}-Clean}, unless otherwise specified.

\subsection{Details of Evaluation Metrics}
\label{sec:appendix_evaluation_metric}
Given a montage and a corresponding natural language needle grounding question, there might be single or multiple target needles on our benchmark.
Conventional temporal grounding datasets~\citep{gao2017tall,lei2021qv} primarily utilize Recall@1 (IoU=0.7) and Recall@1 (IoU=0.5) as their evaluation metrics. 
However, these evaluation metrics are typically designed for single-target grounding questions and are suboptimal for our benchmark, which features a multi-target setting requiring more nuanced evaluation.
Motivated by~\citep{grauman2022ego4d}, we choose Recall@1x, tIoU=0.7, Recall@1x, tIoU=0.5, and Average mAP as the evaluation metrics.
Specifically, the details are listed as follows:

\medskip
\noindent\textbf{Recall@kx, tIoU=m.}
The metric Recall@kx, tIoU=m is specifically designed for temporal grounding tasks with multiple targets, where tIoU denotes the temporal intersection over union, and x presents the number of targets for the given query.
To be specific, Recall@kx, tIoU=m measures the percentage of all the correctly predicted targets that have at least one prediction in the top-kx results of the question with a tIoU exceeding the threshold m.
In this work, we set k=1 and evaluate at two different thresholds, \ie, m=0.7 and m=0.5.

To align with our task setting, we treat a time range as a discrete set of integer timestamps rather than as a continuous interval when calculating tIoU. 
Specifically, given a time range $T_{\mathrm{range}}=[T_{\mathrm{start}},T_{\mathrm{end}}]$, the corresponding integer timestamp set is defined as:
\begin{equation}
\label{eq:s_range}
S_{\mathrm{range}} = \{\, t \in \mathbb{Z} \mid \lfloor T_{\mathrm{start}} \rfloor \le t \le \lceil T_{\mathrm{end}} \rceil \,\}.
\end{equation}
In practice, a ground-truth answer (abbreviated as gt) or a model prediction (abbreviated as pred) may contain multiple time ranges, which can be expressed as:
\begin{equation}
\label{eq:t_c}
T_{c} = \{T_{\mathrm{range}, c}^{1}, T_{\mathrm{range}, c}^{2}, \dots\}, \quad c \in \{\mathrm{gt}, \mathrm{pred}\}
\end{equation}
Based on Eq.~\ref{eq:s_range} and Eq.~\ref{eq:t_c}, the timestamp set of $c$ is defined as:
\begin{equation}
S_{c} = S_{\mathrm{range}, c}^{1} \cup S_{\mathrm{range}, c}^{2} \cup \dots .
\end{equation}

\noindent 
Therefore, the tIoU is calculated as:
\begin{equation}
\mathrm{tIoU} = \frac{\lvert S_{\mathrm{gt}} \cap S_{\mathrm{pred}} \rvert}{\lvert S_{\mathrm{gt}} \cup S_{\mathrm{pred}} \rvert},
\end{equation}
where $\lvert \cdot \rvert$ denotes the cardinality of a set.

\medskip
\noindent\textbf{Average mAP.}
Average mAP is tailored for detection tasks involving multiple targets.
For a given tIoU threshold, mAP (mean average precision) measures how closely the prediction matches the ground-truth answer for each query.
We follow~\citep{grauman2022ego4d} to calculate mAP at five distinct tIoU thresholds \{0.1,0.2,0.3,0.4,0.5\}, and compute the metric Average mAP as the average of these mAP values.

\begin{table}[t!]
\centering
\caption{Refusal rates of closed-source VideoLLMs on \textit{\benchmark{}}.} 
\setlength{\tabcolsep}{0.7mm}{
\begin{tabular}{l|c|c}
\toprule
\textbf{Methods} & \textbf{QA Pairs Answered} & \textbf{Refusal Rate} \\
\midrule
Qwen-VL-Max & 1,967 / 2,053 & 4.19\% \\
GPT-4o & 2,043 / 2,053 & 0.49\% \\
Gemini-1.5-Flash-8B & 2,024 / 2,053 & 1.41\% \\
Gemini-1.5-Flash-002 & 2,014 / 2,053 & 1.90\% \\
Gemini-1.5-Pro-002 & 2,024 / 2,053 & 1.41\% \\
\bottomrule
\end{tabular}
}
\label{tab:refusal_rate}
\end{table}

\subsection{Refusal Rates of Closed-Source Models}
\label{sec:appendix_refusal_rates}
Due to security policies governing the API services, closed-source models may refuse to answer certain questions on \textit{\benchmark{}}. 
As shown in Tab.~\ref{tab:refusal_rate}, the refusal rates of most models are below 2\%, with the exception of Qwen-VL-Max, which exhibits a relatively higher refusal rate (\ie, approximately 4\%). 
Furthermore, we observe that models from the same series may present slight variations in refusal rates, as evidenced by Gemini-1.5-Flash-002 and Gemini-1.5-Pro-002.

\begin{table*}[t]
\centering
\caption{Evaluation results on \textit{\benchmark{}-Full} (short montages). 
\textbf{Avg}: Average Recall@1x, tIoU=0.7 across object and scene needle grounding.
\textbf{Bold}: Best result.
\underline{Underline}: Second-best result.
\textbf{Rank}/{\color{gray}{\textbf{Rank$^\text{C}$}}}: Performance rankings of VideoLLMs on \textit{\benchmark{}-Full}/{\color{gray}\textit{\benchmark{}-Clean}}.
}
\resizebox{\textwidth}{!}{
\begin{tabular}{l|ccc|ccc|ccc}
\toprule

 \multirow{3}{*}{\textbf{Methods}}& \multicolumn{3}{c|}{\textbf{Short Object Needle Grounding}} & \multicolumn{3}{c|}{\textbf{Short Scene Needle Grounding}} & \multirow{3}{*}{\textbf{Avg}} & \multirow{3}{*}{\textbf{Rank}} & \multirow{3}{*}{\color{gray}{\textbf{Rank$^\text{C}$}}} \\
\cmidrule{2-7}
& \textbf{Recall@1x,} & \textbf{Recall@1x,} & \textbf{Average}  & \textbf{Recall@1x,} & \textbf{Recall@1x,} & \textbf{Average} \\
& \textbf{tIoU=0.7}   & \textbf{tIoU=0.5}   & \textbf{mAP}   & \textbf{tIoU=0.7}   & \textbf{tIoU=0.5}   & \textbf{mAP}   \\ 
\midrule
\multicolumn{9}{c}{\textit{\textbf{Open-Source VideoLLMs}}} \\ 
\midrule
LLaVA-Video-7B       & 1.11  & 2.13  & 11.15    & 2.45  & 4.08  & 20.38   & 1.78 & 14 & {\color{gray}{14}} \\ 
LongVA-7B            & 1.62  & 4.46  & 18.45    & 3.36  & 7.54  & 22.75   & 2.49 & 12 & {\color{gray}{13}}\\
VILA1.5-13B             & 0.91  & 2.22  & 4.54     & 3.47  & 6.42  & 14.73   & 2.19 & 13 & {\color{gray}{12}}\\ 
Oryx-34B             & 2.63  & 8.00  & 34.44    & 6.42  & 14.98  & 42.67   & 4.53 & 10 & {\color{gray}{11}}\\ 
MiniCPM v2.6-8B      & 3.65  & 9.52 & 34.87    & 3.87  & 10.70 & 41.93   & 3.76 & 11 & {\color{gray}{10}}\\ 
VILA1.5-40B             & 5.98  & 12.97 & 32.34    & 11.11  & 22.83 & 49.82   & 8.55 & 9 & {\color{gray}{9}}\\
LLaVA-Video-72B      & 9.12  & 18.84 & 43.66    & 17.13 & 28.64 & 52.72   & 13.13 & 7 & {\color{gray}{8}}\\ 
E.T. Chat-3.8B   & 8.13 & 18.09 & 34.50    & 14.65 & 25.51 & 44.97 & 11.39 &  8 & {\color{gray}{7}} \\
LongVU-7B            & 13.27 & 28.88 & 60.81    & 19.67 & 34.05 & 67.75   & 16.47 & 6 & {\color{gray}{6}} \\ \midrule

\multicolumn{9}{c}{\textit{\textbf{Closed-Source VideoLLMs}}} \\ \midrule
Gemini-1.5-Flash-8B  & 37.73 & 46.86 & 70.81    & 56.94 & 62.55 & 81.16   
 & 47.33 & 5 & {\color{gray}{5}}\\ 
Qwen-VL-Max          & 49.18 & 59.77 & 79.62   & 55.96 & 63.11 & 80.95    & 52.57 & 4 & {\color{gray}{4}}\\ 
Gemini-1.5-Flash-002 & 56.59 & 69.57 & 81.23    & \underline{71.53} & 78.47 & 86.90    & 64.06 & 3 & {\color{gray}{3}}\\ 
GPT-4o    & \underline{59.57} & \underline{71.63} & \underline{85.25}  & 71.46 & \underline{78.90} & \underline{87.14} 
& \underline{65.52} & 2 & {\color{gray}{2}}\\
Gemini-1.5-Pro-002   &  \textbf{61.05}  &  \textbf{74.54}  &  \textbf{85.51}    &  \textbf{75.20}  &  \textbf{84.18} &  \textbf{90.48}  & \textbf{68.13} & 1 & {\color{gray}{1}}\\
\bottomrule
\end{tabular}
}
\label{tab:short_exp_noisy}
\end{table*}

\begin{table*}[t]
\centering
\caption{Evaluation results on \textit{\benchmark{}-Full} (medium montages). 
\textbf{Avg}: Average Recall@1x, tIoU=0.7 across object and scene needle grounding.
\textbf{Bold}: Best result.
\underline{Underline}: Second-best result.
\textbf{Rank}/{\color{gray}{\textbf{Rank$^\text{C}$}}}: Performance rankings of VideoLLMs on \textit{\benchmark{}-Full}/{\color{gray}\textit{\benchmark{}-Clean}}.
}
\resizebox{\textwidth}{!}{
\begin{tabular}{l|ccc|ccc|ccc}
\toprule
 \multirow{3}{*}{\textbf{Methods}}& \multicolumn{3}{c|}{\textbf{Medium Object Needle Grounding}} & \multicolumn{3}{c|}{\textbf{Medium Scene Needle Grounding}} & \multirow{3}{*}{\textbf{Avg}} & \multirow{3}{*}{\textbf{Rank}} & \multirow{3}{*}{\color{gray}{\textbf{Rank$^\text{C}$}}}\\
\cmidrule{2-7}
& \textbf{Recall@1x,} & \textbf{Recall@1x,} & \textbf{Average}  & \textbf{Recall@1x,} & \textbf{Recall@1x,} & \textbf{Average} \\
& \textbf{tIoU=0.7}   & \textbf{tIoU=0.5}   & \textbf{mAP}   & \textbf{tIoU=0.7}   & \textbf{tIoU=0.5}   & \textbf{mAP}   \\ 
\midrule
\multicolumn{9}{c}{\textit{\textbf{Open-Source VideoLLMs}}} \\ 
\midrule
MiniCPM v2.6-8B      & 0.26   &  0.79  &    4.19   &  0.53  & 2.37  & 6.12  &  0.40   & 9 & {\color{gray}{9}}\\
LongVU-7B            & 0.79   &  2.12  &   3.64   &  1.32  & 1.84  & 4.19  &  1.06 & 6  & {\color{gray}{8}}   \\ 
E.T. Chat-3.8B    & 0.57   &  0.86  &    2.97   &  0.84  & 3.06  & 5.63  &  0.71   & 8 & {\color{gray}{7}} \\
LLaVA-Video-72B       & 0.53   &  1.85  &    6.32   &  1.05  & 2.89  & 8.83  &  0.79  & 7 & {\color{gray}{6}}   \\
\midrule

\multicolumn{9}{c}{\textit{\textbf{Closed-Source VideoLLMs}}} \\ \midrule
Qwen-VL-Max          & 12.98  & 19.89  & 36.72    & 18.96 & 25.82 & 45.13 & 15.97 & 5 & {\color{gray}{5}} \\
GPT-4o               & 23.28  & 39.15  & 61.96   & 31.05  & 43.42  & 68.04 &  27.17 & 4  & {\color{gray}{4}} \\
Gemini-1.5-Flash-8B  & 33.70 & 41.92 & 55.84   & 46.30 & 54.79 & 71.21 & 40.00 & 3  & {\color{gray}{3}}\\
Gemini-1.5-Flash-002 & \underline{41.64} & \underline{54.25} & \underline{68.34}    & \underline{61.37} & \underline{72.33} & \underline{81.67}  & \underline{55.48} &  2 & {\color{gray}{2}} \\
Gemini-1.5-Pro-002   &  \textbf{48.77}  &  \textbf{59.73}  &  \textbf{70.02}   &  \textbf{68.77} &  \textbf{76.99}  &  \textbf{84.60} & \textbf{58.77} & 1  & {\color{gray}{1}}\\
\bottomrule
\end{tabular}
}
\label{tab:mid_exp_noisy}
\end{table*}

\begin{table*}[t]
\centering
\caption{Evaluation results on \textit{\benchmark{}-Full} (long montages). 
\textbf{Avg}: Average Recall@1x, tIoU=0.7 across object and scene needle grounding.
\textbf{Bold}: Best result.
\underline{Underline}: Second-best result.
\textbf{Rank}/{\color{gray}{\textbf{Rank$^\text{C}$}}}: Performance rankings of VideoLLMs on \textit{\benchmark{}-Full}/{\color{gray}\textit{\benchmark{}-Clean}}.
}
\resizebox{\textwidth}{!}{
\begin{tabular}{l|ccc|ccc|ccc}
\toprule
 \multirow{3}{*}{\textbf{Methods}}& \multicolumn{3}{c|}{\textbf{Long Object Needle Grounding}} & \multicolumn{3}{c|}{\textbf{Long Scene Needle Grounding}} & \multirow{3}{*}{\textbf{Avg}} & \multirow{3}{*}{\textbf{Rank}} & \multirow{3}{*}{\color{gray}{\textbf{Rank$^\text{C}$}}} \\
\cmidrule{2-7}
& \textbf{Recall@1x,} & \textbf{Recall@1x,} & \textbf{Average}  & \textbf{Recall@1x,} & \textbf{Recall@1x,} & \textbf{Average} \\
& \textbf{tIoU=0.7}   & \textbf{tIoU=0.5}   & \textbf{mAP}   & \textbf{tIoU=0.7}   & \textbf{tIoU=0.5}   & \textbf{mAP}   \\ 
\midrule
\multicolumn{9}{c}{\textit{\textbf{Closed-Source VideoLLMs}}} \\ \midrule
Qwen-VL-Max          & 0.48  & 2.38  & 7.55   & 0.95  & 3.81  & 11.73  & 0.72 & 5 & {\color{gray}{5}} \\
GPT-4o               & 2.49  & 6.22  & 22.68     & 3.77  & 8.79  & 29.40 &  3.13 & 4 & {\color{gray}{4}}\\
Gemini-1.5-Flash-8B  & 27.07 & 35.37 & 46.36  & 36.56 & 43.61 & 57.89 & 31.82 & 3 & {\color{gray}{3}}\\
Gemini-1.5-Flash-002 &  \underline{38.43}  &  \underline{53.28}  &  \underline{66.84}  &  \underline{54.63}  &  \underline{65.64}  &  \underline{76.49}  &  \underline{46.53} & 2 & {\color{gray}{2}}\\
Gemini-1.5-Pro-002   &  \textbf{49.78} &  \textbf{58.08}  &  \textbf{67.96}  &  \textbf{62.56}   &  \textbf{68.28}  &  \textbf{79.95}  &  \textbf{56.17} & 1 & {\color{gray}{1}}\\
\bottomrule
\end{tabular}
}
\label{tab:long_exp_noisy}
\end{table*}

\subsection{Comprehensive Experimental Results on \benchmark{}-Full}
\label{sec:appendix_noisy_benchmark}
As described in the main paper, \textit{\benchmark{}-Full} is built upon our automated data generation pipeline, without any manual intervention.
In this section, we present the comprehensive experimental results on the automatically generated \textit{\benchmark{}-Full}.
To ensure a fair comparison, we adopt the same experimental settings as on \textit{\benchmark{}-Clean} to evaluate the performance of advanced VideoLLMs on \textit{\benchmark{}-Full} (using a noisy variant of \textit{\benchmark{}-Clean} that eliminates the manual QA verification). 

As shown in Tab.~\ref{tab:short_exp_noisy}, Tab.~\ref{tab:mid_exp_noisy}, and Tab.~\ref{tab:long_exp_noisy}, we observe that the performance rankings remain consistent between \textit{\benchmark{}-Full} and \textit{\benchmark{}-Clean}.
Specifically, on \textit{\benchmark{}-Full} (short montages), the rankings exhibit only a minor fluctuation of one position among the open-source VideoLLMs, with the corresponding models still achieving comparable performance (\ie, rank 7 and rank 8, rank 10 and rank 11, rank 12 and rank 13).
A similar trend is observed on \textit{\benchmark{}-Full} (medium montages) and \textit{\benchmark{}-Full} (long montages), where the rankings also show only minor fluctuations.
Notably, all closed-source VideoLLMs preserve identical rankings across different montage durations.
These results demonstrate that our automated data generation pipeline can generate high-quality evaluation data sufficient to rank the performance of different models.

{
\subsection{Details of Rank Correlation}
\label{sec:appendix_correlation}
In this section, we present the details of calculating the rank correlation between our \textit{\benchmark{}} and other benchmarks.
As described in the main paper, we denote the Pearson correlation coefficient as $r$ and the p value as $p$.
To ensure a robust calculation, we adopt \textit{\benchmark{}}'s \textit{core} set (short montages) to cover more model performance results.
For our benchmark, we use the performance of Scene Moment Grounding (\ie, Average mAP) for rank correlation calculation.
For other benchmarks, the details are listed as follows: 
\begin{itemize}
    \item E.T. Bench~\cite{etbench}: We use the performance of its temporal video grounding task, which shares a similar task format as our task of ``finding moments in a montage'', where $r_\text{E.T. Bench}$ = 0.9591 and $p_\text{E.T. Bench}$ = 0.0409 $<$ 0.05.
    \item VideoAutoArena~\cite{VideoAutoArena}: We use the performance of ``Win Rates'', where $r_\text{VideoAutoArena} = 0.9055$ and $p_\text{VideoAutoArena}$ = 0.0344 $<$ 0.05.
    \item Chatbot Arena-Vision~\cite{arena}: We use the performance of ``Arena Score'', where $r_\text{Chatbot Arena-Vision}$ = 0.9516 and $p_\text{Chatbot Arena-Vision}$ = 0.0034 $<$ 0.05.
    \item MLVU~\cite{mlvu}: We use the performance of ``M-AVG'' on MLVU-Test. Specifically, our calculation borrows the performance results from~\cite{zhang2024videoinstructiontuningsynthetic}, where$r_\text{MLVU}$ = 0.9983 and $p_\text{MLVU}$ = 0.0368 $<$ 0.05.
    \item EgoSchema~\cite{egoschema}: We use the performance on EgoSchema-test, where $r_\text{EgoSchema}$ = 0.8619 and $p_\text{EgoSchema}$ = 0.0126 $<$ 0.05 .
    \item Video-MME~\cite{videomme}: We use the performance of ``Overall w/o subs'', where $r_\text{Video-MME}$ = 0.7771 and $p_\text{Video-MME}$ = 0.0233 $<$ 0.05.
\end{itemize}
}

\section{Details of Our Automated Data Generation Pipeline}
\label{sec:appendix_pipeline}

\subsection{Details of Video Representation Extraction}
\label{sec:appendix_video_representation}
In this section, we provide the implementation details of extracting scene tables and object tables of the proposed comprehensive video representation.
For scene tables, we first employ \texttt{PySceneDetect}\footnote{\url{https://www.scenedetect.com/}} to segment a video into multiple scenes (\i, multiple distinct, exclusive short video clips). 
Then, we utilize RAM++ ~\citep{huang2023opensetimagetaggingmultigrained} to tag all existing objects within each scene. 
Subsequently, we construct the object tables for these tagged objects and establish object table links from the scene table to these object tables.
To be specific, we use UniRef~\citep{DBLP:conf/emnlp/ZhengKJWW22} to generate the referring expression for the object, and employ DEVA~\citep{cheng2023deva} to obtain the tracking sequence (\ie, a sequence of bounding boxes) of the object.

\subsection{Details of NeMo Data Generation}
During the stage of NeMo Data Generation, we first select the target scene containing salient visual objects based on the extracted video representation. To be specific, the target scene must satisfy the following requirements: 
(1) Its duration is longer than five seconds. 
(2) It should contain at least one visually prominent object that is visible for at least four seconds, with an average visible area between 5\% and 25\% of the frame.

Tab.~\ref{tab:data_generation_prompts} and Tab.~\ref{tab:data_generation_prompts2} present the prompts for needle grounding QA generation and self-verification used in our pipeline. 
To facilitate the joint automated generation of QA pairs for object needle grounding and scene needle grounding, we first extract frames from the target scene, ensuring that the target object is encircled with a bounding box in each frame.
The extracted frames are then fed into a frozen multimodal model (\ie, GPT-4o) via visual prompting to generate the corresponding object description, scene description, and a scene needle grounding question. 
As shown in Tab.~\ref{tab:object_localization_qa_template}, we apply the predefined templates to the object descriptions to automatically generate the corresponding object needle grounding questions.
Furthermore, we retain only the QA pairs that successfully pass the self-verification process conducted by the same frozen VideoLLM, thereby mitigating the inaccuracies and hallucinations introduced by the annotation models.

\subsection{Progressive Data Expansion: From Short to Long}
In this section, we provide the details of extending the short montages into medium montages and long montages.

\medskip
\noindent\textbf{From Short Montage to Medium Montage.}
Given a short montage with a target scene, similar to Stage 2 (see the main paper for more details), we synthesize a medium montage (between 2.5 and 15 minutes) by seamlessly integrating the target scene with more ``negative'' short video clips from the same raw video, while preserving a consistent shooting style.

\medskip
\noindent\textbf{From Short Montage to Long Montage.}
For constructing a long montage (more than 15 minutes), we consider the ``negative'' short video clips not only from the same raw video but also from other raw videos within the same video program, ensuring consistency in shooting style and similarity in video content.
Moreover, given the high API cost of evaluating closed-source VideoLLMs on long montages, we allow the construction of long montages with multiple target scenes and multiple target objects, thereby supporting multiple grounding QA pairs in a single long montage.
To be specific, each long montage has an average of 8.11 QA pairs.

\subsection{Progressive Data Expansion: From Single to Multi}
For constructing multi-needle QA pairs, we select only the single-needle QA pairs where the durations of the target scene and the target object (within the target scene) exceed six seconds.
Then, we segment the target scene into multiple video clips, ensuring that each segmented video clip lasts for more than two seconds. 
Further, we reconstruct a new multi-needle montage by removing the target scene from the original montage and then randomly inserting those segmented video clips into the remaining content.

\subsection{Manual Verification}
\label{sec:appendix_manual_verification}
The raw QA pairs of \textit{\benchmark{}-Full} and \textit{\benchmark{}-Clean} are both automatically generated by the proposed data generation pipeline.
To facilitate a robust, precise evaluation of VideoLLMs, we engage human annotators to conduct manual verification on \textit{\benchmark{}-Clean}: (1) refining the generated question if it misaligns with the target object/scene, and (2) refining the generated answer if there are incorrect temporal boundaries, missing target needles, or the presence of unnecessary target needles. 
Specifically, the manual verification is employed after both Stage 2 and Stage 3 of our pipeline. 

{
\subsection{Analysis on Self-Verification}
\label{sec:appendix_self_verification}
In this part, we provide a detailed analysis of our self-verification strategy.
As described in Sec.~\ref{sec:pipeline}, our automated data generation pipeline adopts a self-verification strategy to mitigate potential inaccuracies and hallucinations introduced by annotation models.
Specifically, the rejection rate of the self-verification step is $\sim$32.34\%, where we observe that the rejected cases mainly involve: 
(1) ambiguous object- or scene-level needle grounding questions (\eg, the generated object description cannot uniquely identify the target object, or the generated scene description corresponds to multiple candidate scenes);
(2) inaccurate object-level needle grounding questions (\eg, the generated object description contains incorrect visual attributes, such as colors, categories, or fine-grained appearance details);
(3) inconsistent scene-level needle grounding questions (\eg, the generated scene description is not well aligned with the visual content of the target moment);
(4) subtitle-dependent scene descriptions (\eg, the generated scene description relies on information that can only be obtained from subtitles rather than visual evidence);
and (5) distractor-matching cases (\eg, non-target objects or scenes accidentally satisfy the generated object or scene description).
These cases demonstrate the effectiveness of self-verification in mitigating potential inaccuracies and hallucinations introduced by annotation models, thereby improving the quality of automated data generation.
}

{
\section{Additional Discussions}
\label{sec:appendix_discussion}

\subsection{Comparisons in Needle Design}
In this section, we discuss the difference between the needle used in the standard ``needle in a haystack'' (NIAH) and our \textit{\task{}}.

\medskip
\noindent\textbf{Needle design in NIAH.} 
As described in Sec. 1, the standard NIAH is implemented by placing a random fact or statement (\ie, the ``irrelevant needles'') in the middle of a long context window (\ie, the haystack), without any semantic connection.
Moreover, as discussed in Sec. 1 and Sec. 2.2, prior works in video-language benchmarks (\eg, VNBench) also follow a similar design paradigm.
For example, Fig.2 presents that VNBench inserts static fruit images (\ie, ``video-irrelevant needles'' with notable visual and stylistic differences from the original video content) into a video haystack, inheriting the limitations of the original NIAH. 

\medskip
\noindent\textbf{Needle design in \task{}.}
Motivated by the montage synthesis technology widely used in real-life applications, 
our \textit{} is specifically designed to mitigate the issue of ``irrelevant needles''. We clarify the uniqueness of our needle design as follows:
(1) As in Sec. 1, Sec. 2.2, and Sec. 3, we use ``loosely related clips'' from the same video source to form a montage. The ``needles'' can be considered ``video-relevant needles'', as the target needles share similar shooting styles, entities, and backgrounds with the surrounding montage clips, while still maintaining a coherent theme across the montage. 
(2) Our visualizations further support that the surrounding montage clips share similar visual semantics to the target needles. 
For example, in Fig. 6 (middle), both the surrounding montage clips and the target needles depict the daily activities of the same person.
Similarly, in Fig. 8 (left), both the surrounding montage clips and the target needles present the behavior of pandas.
These examples demonstrate that the ``needles'' in our task are ``video-relevant'' (\ie, the context around the needles remains semantically and visually connected, unlike the original NIAH), consistent with our task design.


\subsection{Montages vs. Natural Long-Form Videos}
\label{sec:appendix_montage_vs_real}
In this section, we provide a detailed discussion of the necessity of using synthetic montages for our benchmark construction, instead of using natural long-form videos.

\medskip
\noindent\textbf{Considerations in manual annotation cost and automated annotation feasibility for \task{}.}
(1) Manual annotation cost: As described in Sec. 1 and Sec. 3.2, we would like to point out that ``annotating videos, especially those of long duration, demands substantial manual effort, restricting scalability''.
For example, we have detailed the annotation cost of the conventional manual annotation process (ManualAnno) in Sec. 4 and Appendix B.
Specifically, ManualAnno requires three manual steps (\ie, question construction phase, answer construction phase, and cleaning phase) to ensure high-quality data generation for temporal grounding tasks.
Moreover, we have conducted both theoretical and practical analyses to demonstrate the high cost of ManualAnno in Sec. 4.1 and Sec. 4.2.

(2) Automated annotation feasibility: To generate data for our \textit{\task{}} task while reducing manual annotation cost, an intuitive solution is to directly employ strong annotation models (\eg, GPT-4o) on the natural long-form videos.
However, during the development of our benchmark, we observed that this solution is infeasible for annotating long videos (especially for hour-long videos) due to the limited context length of these annotation models.
For example, annotation models often fail to extract comprehensive object-level and scene-level information (\eg, detailed attributes of all objects throughout the video) in long videos.
To address this issue, in Sec. 3.2, we proposed a video representation containing comprehensive scene- and object-level features for montage clips, and introduced an automated data generation pipeline specifically designed for our task of ``needle in a montage''.

\medskip
\noindent\textbf{Considerations in scalability and annotation quality for our benchmark construction.}
(1) Benchmark scalability: As discussed above, manually annotating natural long-form videos incurs prohibitive manual annotation cost, making it difficult to scale. 
As mentioned in Sec. 2.2, ``these benchmarks still heavily rely on manual annotation, which is both labor-intensive and time-consuming, thereby hindering the scalability of benchmark construction''.
At the same time, we would like to emphasize that cost-effectiveness and scalability are critical for enabling continuous data integration and evaluation, ``thereby preventing potential data contamination in large model evaluations'' (stated in Sec. 3.2.2). 
Therefore, we choose to construct the \textit{\benchmark{}} using our proposed scalable, automated pipeline featuring montage synthesis.

(2) High-quality data with low annotation cost: Another important consideration in constructing a benchmark is data quality. 
As described in Sec 4, we have conducted ``a comparative analysis of our data generation pipeline in terms of time efficiency and data quality''. 
The results present that our automated pipeline ``can automatically generate near-human, high-quality evaluation data'', while demonstrating ``significant time efficiency''.
Specifically, compared with ManualAnno, our automated pipeline achieves comparable annotation accuracy (Sec. 4.2 and Tab. 3), while yielding a time reduction ratio of $\sim$0.78, \ie, $\sim$4.5x reduction in time (Sec. 4.1).
Moreover, we observe that model rankings remain consistent between \textit{\benchmark{}-Clean} and \textit{\benchmark{}-Full} in Sec. 5.2.
These results further demonstrate the effectiveness of our automated pipeline with montage synthesis.

\subsection{Clarification of Retrieval-Style Long-Context Recall in Our Work.}
In this section, we clarify that the term ``long-context recall'' can be interpreted at different levels of capability.
For example, a more complicated form of long-context video understanding may require models to connect information across distant temporal points, such as recalling a character's earlier statement to understand a later action.
This setting may also involve cross-segment reasoning or information integration, and can be viewed as a form of \textbf{cross-segment long-context reasoning}.
By contrast, our \textit{\task{}} task focuses on \textbf{retrieval-style long-context recall}.
Specifically, given a needle grounding question and a synthesized montage containing a long visual sequence, the model is required to identify all query-relevant target needles and localize their temporal extents, as described in Sec.~\ref{sec:task_suite}.
This formulation is closely related to conventional temporal grounding in long videos and information retrieval in long-context documents, where the key challenge lies in accurately retrieving the relevant evidence from a long context.
Under this definition, although the ``long-context recall'' evaluated in our task does not necessarily require cross-segment reasoning or connecting information across distant temporal points, it remains a fundamental prerequisite for higher-level cross-segment reasoning, as such reasoning cannot be reliably performed without first identifying the relevant evidence from a long context.

\subsection{Additional Discussion on the Shortcut Baseline}
In this section, we discuss that needle-in-a-haystack benchmarks designed to assess long-context recall can often be partially approached without holistic long-context processing.
Specifically, a model could theoretically segment the haystack and check each segment independently, thereby obtaining a potential shortcut solution, although this is not the standard evaluation setting.
For example, in the standard needle-in-a-haystack test, given a multi-needle question (\eg, ``What are the secret ingredients needed to build the perfect pizza?''), such a shortcut baseline can be implemented by first segmenting the long-context haystack (\eg, a long-context document) into short sentences, and then performing needle detection over each sentence sequentially.
The retrieved needles (\eg, ``Figs are one of the secret ingredients ...'' and ``Prosciutto is one of the secret ingredients ...'') can be subsequently aggregated to answer the question.
Similarly, the shortcut baseline in our task aligns with this divide-and-conquer strategy, which first segments the input into shorter video clips and then performs temporal grounding sequentially.

However, the proposed shortcut baseline is not the standard evaluation setting of our benchmark, and does not invalidate the claim that our benchmark evaluates long-context recall.
Specifically, our benchmark is designed as an end-to-end, single-turn evaluation to assess VideoLLMs' inherent capabilities (including both retrieval-style long-context recall and temporal grounding) in retrieving target needles from a long-context montage and localizing their temporal extents.
During the development of our benchmark, we observed that prior studies rarely discussed the effectiveness of such a shortcut baseline in needle-in-a-haystack tests.
To bridge this gap, we introduce the shortcut baseline and provide an empirical analysis in Sec. 5.3.
Moreover, the experimental results of the shortcut baseline offer useful insights for future model development.
For example, while the shortcut baseline yields partial improvement over the base model across different montage durations, it still significantly lags behind advanced closed-source VideoLLMs.
These results underscore the need to enhance the inherent long-context recall and temporal grounding capabilities of VideoLLMs.

\subsection{Analysis of Splice Points in Montages}
In this section, we provide a diagnostic analysis of whether VideoLLMs can detect splice points in synthesized montages.
We have conducted an additional diagnostic experiment on \textit{\benchmark{}-Clean} (medium montages) using Qwen2.5-VL-72B, the best-performing open-source model in our experiments.
Considering that both real-world videos (\eg, professionally produced TV shows) and our synthesized montages commonly contain scene transitions, we evaluate whether the model can infer the underlying splice structure by asking it to predict the number of concatenated clips in each montage.
We adopt accuracy as the evaluation metric, which is computed by exact matching between the predicted and ground-truth numbers of clips.
The experimental result reveals that Qwen2.5-VL-72B achieves 0\% accuracy, indicating that even the best open-source model struggles to detect and recover the montage structure.
This finding suggests that, although montage construction naturally introduces clip boundaries (or splice points), these boundaries do not provide an easy or reliable shortcut for the evaluated models.
Additionally, we would like to point out that this observation is also consistent with our \textit{\task{}} task design, where montages are synthesized by  
``loosely related clips'' from the same video source, and the target needles share similar shooting styles, entities, and backgrounds with the surrounding montage clips, thereby maintaining a coherent theme across the montage.

\subsection{Additional Experiments of Needle Relocation}
\label{sec:appendix_needle_relocation}
In this section, to investigate the effects of needle relocation on the model performance, we have conducted a paired relocation experiment on our \textit{\benchmark{}} (\ie, \textit{\benchmark{}-Original} vs. \textit{\benchmark{}-Relocation}).
Following the reviewer's suggestion, \textit{\benchmark{}-Relocation} is reconstructed from \textit{\benchmark{}-Original}, where ``the same needle is placed in different positions within the same montage''.
Specifically, for each target needle within a montage, we preserve its video content and duration, and only relocate it to a different temporal position within the same montage.
To provide a concise and reliable diagnostic analysis, the experiments are conducted using two specialist VideoLLMs for temporal grounding (\ie, Time-R1-3B and Time-R1-7B), as shown in Tab.~\ref{tab:needle_position_bias_medium_timer1}, Tab.~\ref{tab:relocate_compare_mid_avg}, and Tab.~\ref{tab:relocate_compare_mid_merged}.
The detailed analyses are as follows:

\medskip
\noindent\textbf{Models' inherent positional bias.}
Before analyzing the needle relocation results, we first examine model performance across different needle positions on \textit{\benchmark{}-Original}.
As shown in Tab.~\ref{tab:needle_position_bias_medium_timer1}, both evaluated models achieve their best Avg performance when the needle appears in the beginning range, indicating their positional biases towards the beginning range.
These observations are also consistent with the findings in Sec. 5.3, where we show that positional bias exists but is model-dependent across different VideoLLMs.

\medskip
\noindent\textbf{Aggregated-level results of needle relocation on models' temporal grounding performance.}
To examine whether the needle relocation affects models' temporal grounding performance, we analyze the aggregated results from two complementary perspectives, based on the results in Tab.~\ref{tab:needle_position_bias_medium_timer1} and Tab.~\ref{tab:relocate_compare_mid_avg}.
Specifically, although the models exhibit notable performance fluctuations caused by positional biases (Tab.~\ref{tab:needle_position_bias_medium_timer1}), the aggregated results show that both evaluated models maintain broadly comparable temporal grounding performance after relocation (Tab.~\ref{tab:relocate_compare_mid_avg}).
This aggregated-level comparison suggests that model performance is not solely determined by a fixed-position shortcut (the fine-grained transition-level results will be discussed in the next part to analyze how specific source-to-target needle position changes affect model performance).
For aggregated-level results:

(1) Stronger temporal grounding model (\ie, Time-R1-7B): 
In Tab.~\ref{tab:needle_position_bias_medium_timer1}, we observe that Time-R1-7B (beginning range) significantly outperforms Time-R1-7B (middle range) and Time-R1-7B (end range) by +5.10\% and +8.99\% Recall@1x, tIoU=0.7, respectively, on object needle grounding. 
We then compare the aggregated performance difference between \textit{\benchmark{}-Relocation} and \textit{\benchmark{}-Original} in Tab.~\ref{tab:relocate_compare_mid_avg}, where Time-R1-7B only exhibits slight performance fluctuations on medium object needle grounding (\ie, $\boldsymbol{\Delta}$=+0.55\%).
These results suggest that the stronger temporal grounding model, despite its inherent positional bias, can still maintain comparable temporal grounding performance under paired relocation.

(2) Weaker temporal grounding model (\ie, Time-R1-3B): Tab.~\ref{tab:needle_position_bias_medium_timer1} presents that Time-R1-3B (beginning range) similarly outperforms Time-R1-3B (middle range; zero performance) and Time-R1-3B (end range; zero performance) by +0.63\% and +0.63\% Recall@1x, tIoU=0.7, respectively, on object needle grounding. 
In Tab.~\ref{tab:relocate_compare_mid_avg}, due to the relatively weak temporal grounding capability of Time-R1-3B (\ie, yielding near-zero Recall@1x, tIoU=0.7 performance on both \textit{\benchmark{}-Original} and \textit{\benchmark{}-Relocation}), we conduct analysis using the Average mAP metric instead.
Specifically, we observe that Time-R1-3B also exhibits slight performance fluctuations on medium object needle grounding, where the Average mAP increases from 4.27\% to 4.83\%. 
These results indicate that the weaker temporal grounding model, despite its inherent positional bias and limited high-precision temporal grounding performance, still shows weak but non-zero temporal grounding signals under the paired relocation setting.

\medskip
\noindent\textbf{Fine-grained transition-level results of needle relocation on models' temporal grounding performance.}
Tab.~\ref{tab:relocate_compare_mid_merged} further presents the fine-grained transition-level results of Tab.~\ref{tab:relocate_compare_mid_avg}.
For \textit{\benchmark{}-Relocation}, each row corresponds to a specific relocation transition from the original temporal range (\textbf{From}) to the target temporal range after relocation (\textbf{To}).
For each transition, the \textit{\benchmark{}-Relocation} scores are computed on the relocated examples, while the \textit{\benchmark{}-Original} scores are re-computed on the corresponding original examples from the same paired subset.
Therefore, \textit{\benchmark{}-Original} scores provide the original, before-relocation temporal grounding performance for the same needles, enabling a paired comparison with their after-relocation performance.
This ensures that each comparison is conducted under matched needle identity and montage context.
These transition-level results provide a more fine-grained view of how target needles' temporal position affects model performance:

(1) Stronger temporal grounding model (\ie, Time-R1-7B): 
In Tab.~\ref{tab:relocate_compare_mid_merged}, we observe that relocating needles from later temporal ranges (\ie, 33--67\% and 67--100\%) to the beginning range (\ie, 0--33\%) often improves performance.
For example, on object needle grounding, Recall@1x at tIoU=0.7 increases by +11.67\% for the 33--67\% $\rightarrow$ 0--33\% transition and by +22.22\% for the 67--100\% $\rightarrow$ 0--33\% transition.
Similarly, on scene needle grounding, the corresponding performance gains are +17.65\% and +33.33\%, respectively.
In contrast, relocating needles from the beginning range to the middle range leads to performance drops, with $\boldsymbol{\Delta}$=-6.77\% for object needles and $\boldsymbol{\Delta}$=-11.35\% for scene needles.

(2) Weaker temporal grounding model (\ie, Time-R1-3B): 
Similarly, we use the Average mAP metric to analyze the model behavior of Time-R1-3B due to its near-zero performance on high-precision temporal grounding.
In Tab.~\ref{tab:relocate_compare_mid_merged}, we observe that relocating needles from later temporal ranges to the beginning range improves temporal grounding performance.
For example, on object needle grounding, Average mAP increases by +10.33\% for the 33--67\% $\rightarrow$ 0--33\% transition and by +6.67\% for the 67--100\% $\rightarrow$ 0--33\% transition.
Similarly, on scene needle grounding, the corresponding performance gains are +16.76\% and +17.78\%, respectively.
In contrast, relocating needles from the beginning range to the middle range leads to performance drops, with $\boldsymbol{\Delta}$=-5.85\% for object needles and $\boldsymbol{\Delta}$=-9.53\% for scene needles.

These transition-level results support a nuanced conclusion: the paired relocation setting shows evidence of temporal grounding beyond a fixed-position shortcut, especially for the stronger model, while also revealing that model performance is not fully invariant to the temporal position of the needle.

Overall, the above results indicate that ``genuine visual grounding'' (\ie, temporal grounding capability) and positional bias coexist.
On the one hand, the aggregated-level relocation results show that models can still retrieve and temporally ground the relocated needles with broadly comparable performance, suggesting evidence of temporal grounding beyond a fixed-position shortcut.
On the other hand, the transition-level relocation results reveal clear position-dependent variations, indicating that model performance is not fully invariant to the temporal position of the needles.
}

\begin{table*}[ht]
\centering
\caption{{Model performance across various needle positions on \textit{\benchmark{}-Clean} (medium montages; \textit{\benchmark{}-Original}). 
\textbf{Avg}: Average Recall@1x, tIoU=0.7 across object and scene needle grounding. \textbf{Bold}: Best non-zero result across beginning range (0--33\%), middle range (33--67\%), and end range (67--100\%).}}
\renewcommand\arraystretch{0.92}
\resizebox{\textwidth}{!}{
\setlength{\tabcolsep}{1.5mm}{
\begin{tabular}{l|c|ccc|ccc|c}
\toprule
\multirow{3}{*}{\textbf{Methods}} & \multirow{3}{*}{\textbf{Position}} & \multicolumn{3}{c|}{\textbf{Medium Object Needle Grounding}} & \multicolumn{3}{c|}{\textbf{Medium Scene Needle Grounding}} & \multirow{3}{*}{\textbf{Avg}} \\
\cmidrule{3-8}
& & \textbf{Recall@1x,} & \textbf{Recall@1x,} & \textbf{Average} & \textbf{Recall@1x,} & \textbf{Recall@1x,} & \textbf{Average} \\
& & \textbf{tIoU=0.7} & \textbf{tIoU=0.5} & \textbf{mAP} & \textbf{tIoU=0.7} & \textbf{tIoU=0.5} & \textbf{mAP} \\

 \midrule
\multirow{3}{*}{Time-R1-3B} & 0--33\% & \textbf{0.63} & 3.14 & 7.58 & \textbf{1.96} & 5.88 & 11.28 & \textbf{1.29} \\
 & 33--67\% & 0.00 & 0.00 & 1.69 & 0.00 & 0.00 & 2.40 & 0.00 \\
 & 67--100\% & 0.00 & 0.00 & 1.69 & 0.00 & 0.00 & 2.64 & 0.00 \\
 
\midrule
\multirow{3}{*}{Time-R1-7B} & 0--33\% & \textbf{11.95} & 22.01 & 33.31 & \textbf{16.34} & 27.45 & 41.07 & \textbf{14.14} \\
 & 33--67\% & 6.85 & 18.49 & 28.30 & 12.60 & 25.20 & 42.96 & 9.72 \\
 & 67--100\% & 2.96 & 6.67 & 20.97 & 3.20 & 11.20 & 25.60 & 3.08 \\

\bottomrule
\end{tabular}
}
}
\label{tab:needle_position_bias_medium_timer1}
\end{table*}

\begin{table*}[ht]
\centering
\caption{{Aggregated-level results of needle relocation on the model performance on \textit{\benchmark{}-Clean} (medium montages; \textit{\benchmark{}-Original} vs. \textit{\benchmark{}-Relocation}). $\boldsymbol{\Delta}$ denotes the performance difference between \textit{\benchmark{}-Relocation} and \textit{\benchmark{}-Original} (measured in Average mAP for Time-R1-3B, and Recall@1x, tIoU=0.7 for Time-R1-7B).}}
\renewcommand{\arraystretch}{0.92}
\resizebox{\textwidth}{!}{
\setlength{\tabcolsep}{1.5mm}{
\begin{tabular}{l|c|ccc|ccc|c}
\toprule
\multirow{3}{*}{\textbf{Methods}} & \multirow{3}{*}{\begin{tabular}{@{}c@{}}\textbf{Needle} \\ \textbf{Type}\end{tabular}} &  \multicolumn{3}{c|}{\textbf{\textit{\benchmark{}-Original}}} & \multicolumn{3}{c|}{\textbf{\textit{\benchmark{}-Relocation}}} & \multirow{3}{*}{$\boldsymbol{\Delta}$} \\
\cmidrule{3-8}
& &  \textbf{Recall@1x,} & \textbf{Recall@1x,} & \textbf{Average} & \textbf{Recall@1x,} & \textbf{Recall@1x,} & \textbf{Average} & \\
& & \textbf{tIoU=0.7} & \textbf{tIoU=0.5} & \textbf{mAP} & \textbf{tIoU=0.7} & \textbf{tIoU=0.5} & \textbf{mAP} & \\
\midrule

\multirow{2}{*}{Time-R1-3B} & Object  & 0.45 & 1.36 & 4.27 & 0.00 & 0.23 & 4.83 & +0.56 \\
 & Scene  & 0.74 & 2.47 & 5.81 & 0.25 & 0.99 & 8.70 & +2.89 \\
 
 \midrule
\multirow{2}{*}{Time-R1-7B} & Object  & 7.95 & 15.91 & 35.33 & 8.50 & 15.24 & 37.55 & +0.55 \\
 & Scene  & 11.88 & 20.54 & 44.95 & 12.90 & 25.06 & 50.36 & +1.02 \\

\bottomrule
\end{tabular}
}
}
\label{tab:relocate_compare_mid_avg}
\end{table*}

\begin{table*}[ht]
\centering
\caption{{Fine-grained transition-level results of needle relocation on \textit{\benchmark{}-Clean} (medium montages; \textit{\benchmark{}-Original} vs. \textit{\benchmark{}-Relocation}). 
For \textit{\benchmark{}-Relocation}, each row corresponds to a specific relocation transition from the original temporal range (\textbf{From}) to the target temporal range after relocation (\textbf{To}).
For each transition, the \textit{\benchmark{}-Relocation} scores are computed on the relocated examples, while the \textit{\benchmark{}-Original} scores are re-computed on the corresponding original examples from the same paired subset.
Therefore, \textit{\benchmark{}-Original} scores provide the original, before-relocation temporal grounding performance for the same needles, enabling a paired comparison with their after-relocation performance.
$\boldsymbol{\Delta}$ denotes the performance difference between \textit{\benchmark{}-Relocation} and \textit{\benchmark{}-Original} (measured in Average mAP for Time-R1-3B, and Recall@1x, tIoU=0.7 for Time-R1-7B).}}
\renewcommand{\arraystretch}{0.88}
\resizebox{\textwidth}{!}{
\setlength{\tabcolsep}{1.2mm}{
\begin{tabular}{l|c|cc|ccc|ccc|c}
\toprule
\multirow{3}{*}{\textbf{Methods}} & \multirow{3}{*}{\begin{tabular}{@{}c@{}}\textbf{Needle} \\ \textbf{Type}\end{tabular}} & \multicolumn{2}{c|}{\textbf{Needle Position}}  & \multicolumn{3}{c|}{\textbf{\textit{\benchmark{}-Original}}} & \multicolumn{3}{c|}{\textbf{\textit{\benchmark{}-Relocation}}} & \multirow{3}{*}{$\boldsymbol{\Delta}$} \\
\cmidrule{3-10}
& & \multirow{2}{*}{\textbf{From}} & \multirow{2}{*}{\textbf{To}} & \textbf{Recall@1x,} & \textbf{Recall@1x,} & \textbf{Average} & \textbf{Recall@1x,} & \textbf{Recall@1x,} & \textbf{Average} & \\
& & & & \textbf{tIoU=0.7} & \textbf{tIoU=0.5} & \textbf{mAP} & \textbf{tIoU=0.7} & \textbf{tIoU=0.5} & \textbf{mAP} & \\
\midrule

\multirow{18}{*}{Time-R1-3B} & \multirow{9}{*}{Object} & \multirow{3}{*}{0--33\%}  & 0--33\% & 3.45 & 3.45 & 6.21 & 0.00 & 0.00 & 4.29 & -1.92 \\
 & & & 33--67\% & 0.00 & 2.26 & 7.97 & 0.00 & 0.00 & 2.12 & -5.85 \\
 & & & 67--100\% & 0.00 & 0.94 & 3.58 & 0.00 & 0.00 & 4.15 & +0.57 \\
 \cmidrule{3-11}
 & & \multirow{3}{*}{33--67\%}  & 0--33\% & 0.00 & 0.00 & 1.00 & 0.00 & 0.00 & 11.33 & +10.33 \\
 & & & 33--67\% & 0.00 & 0.00 & 0.00 & 0.00 & 0.00 & 3.16 & +3.16 \\
 & & & 67--100\% & 0.00 & 0.00 & 0.00 & 0.00 & 0.00 & 0.54 & +0.54 \\
 \cmidrule{3-11}
 & & \multirow{3}{*}{67--100\%}  & 0--33\% & 0.00 & 0.00 & 0.00 & 0.00 & 0.00 & 6.67 & +6.67 \\
 & & & 33--67\% & 0.00 & 0.00 & 0.00 & 0.00 & 2.78 & 3.89 & +3.89 \\
 & & & 67--100\% & 0.00 & 0.00 & 0.00 & 0.00 & 0.00 & 0.00 & 0.00 \\
\cmidrule{2-11}
 & \multirow{9}{*}{Scene} & \multirow{3}{*}{0--33\%} & 0--33\% & 0.00 & 0.00 & 0.71 & 0.00 & 0.00 & 14.29 & +13.58 \\
 & & & 33--67\% & 1.74 & 5.22 & 14.09 & 0.00 & 0.88 & 4.56 & -9.53 \\
 & & & 67--100\% & 1.05 & 3.16 & 6.32 & 0.00 & 1.05 & 5.05 & -1.27 \\
 \cmidrule{3-11}
 & & \multirow{3}{*}{33--67\%} & 0--33\% & 0.00 & 0.00 & 0.59 & 0.00 & 1.47 & 17.35 & +16.76 \\
 & & & 33--67\% & 0.00 & 0.00 & 6.67 & 0.00 & 0.00 & 0.00 & -6.67 \\
 & & & 67--100\% & 0.00 & 0.00 & 0.61 & 0.00 & 0.00 & 0.00 & -0.61 \\
 \cmidrule{3-11}
 & & \multirow{3}{*}{67--100\%}  & 0--33\% & 0.00 & 0.00 & 0.00 & 0.00 & 0.00 & 17.78 & +17.78 \\
 & & & 33--67\% & 0.00 & 0.00 & 0.00 & 2.56 & 2.56 & 5.64 & +5.64 \\
 & & & 67--100\% & 0.00 & 0.00 & 0.00 & 0.00 & 0.00 & 0.00 & 0.00 \\
\midrule

\multirow{18}{*}{Time-R1-7B} & \multirow{9}{*}{Object} & \multirow{3}{*}{0--33\%}  & 0--33\% & 6.90 & 13.79 & 27.59 & 3.57 & 7.14 & 27.86 & -3.33 \\
 & & & 33--67\% & 12.88 & 24.24 & 33.94 & 6.11 & 12.21 & 20.00 & -6.77 \\
 & & & 67--100\% & 1.89 & 2.83 & 10.57 & 2.83 & 3.77 & 14.15 & +0.94 \\
 \cmidrule{3-11}
 & & \multirow{3}{*}{33--67\%}  & 0--33\% & 13.33 & 35.00 & 52.33 & 25.00 & 45.00 & 60.33 & +11.67 \\
 & & & 33--67\% & 0.00 & 0.00 & 10.53 & 0.00 & 0.00 & 9.47 & 0.00 \\
 & & & 67--100\% & 0.00 & 2.70 & 12.43 & 0.00 & 0.00 & 10.81 & 0.00 \\
 \cmidrule{3-11}
 & & \multirow{3}{*}{67--100\%} & 0--33\% & 0.00 & 11.11 & 40.00 & 22.22 & 44.44 & 66.67 & +22.22 \\
 & & & 33--67\% & 11.76 & 23.53 & 41.18 & 23.53 & 38.24 & 52.94 & +11.76 \\
 & & & 67--100\% & 0.00 & 0.00 & 13.33 & 0.00 & 0.00 & 8.89 & 0.00 \\
\cmidrule{2-11}
 & \multirow{9}{*}{Scene} & \multirow{3}{*}{0--33\%} & 0--33\% & 14.29 & 17.86 & 37.86 & 7.14 & 21.43 & 42.86 & -7.14 \\
 & & & 33--67\% & 17.54 & 29.82 & 43.16 & 6.19 & 16.81 & 25.13 & -11.35 \\
 & & & 67--100\% & 1.05 & 3.16 & 14.11 & 2.11 & 5.26 & 12.63 & +1.05 \\
 \cmidrule{3-11}
 & & \multirow{3}{*}{33--67\%} & 0--33\% & 22.06 & 42.65 & 67.94 & 39.71 & 64.71 & 81.47 & +17.65 \\
 & & & 33--67\% & 11.11 & 22.22 & 24.44 & 0.00 & 0.00 & 17.78 & -11.11 \\
 & & & 67--100\% & 0.00 & 3.03 & 9.09 & 0.00 & 3.03 & 12.73 & 0.00 \\
 \cmidrule{3-11}
 & & \multirow{3}{*}{67--100\%} & 0--33\% & 0.00 & 11.11 & 40.00 & 33.33 & 66.67 & 97.78 & +33.33 \\
 & & & 33--67\% & 10.26 & 30.77 & 46.67 & 28.21 & 51.28 & 64.62 & +17.95 \\
 & & & 67--100\% & 0.00 & 0.00 & 22.22 & 0.00 & 0.00 & 11.11 & 0.00 \\
\bottomrule
\end{tabular}
}
}
\label{tab:relocate_compare_mid_merged}
\end{table*}

\begin{table*}[t!]
\centering
\caption{Prompts for needle grounding QA generation.}
\setlength{\tabcolsep}{0.7mm}
{
\begin{tabular}{p{\textwidth}}
\hline
\toprule
\textbf{Needle Grounding QA Generation}: \\
\midrule
Please help me generate a scene description, a question based on the scene description for the scene needle grounding problem, and an object description for building questions about object needle grounding.\\
There are \{frame num\} frames from a video. Based on the object in the red bounding box, the category of the object, the noisy description of the object, and the noisy subtitles of the video which may not reflect the visual information of this frame, generate the whole scene description, the corresponding question for scene needle grounding problem and the concrete object description that can help people quickly identify this object.\\
Object Category: \{category\}\\
Noisy Object Description: \{desc\}\\
Subtitle: \{subtitle text\}\\
The provided object should be concrete, specific, and identifiable. If the current object belongs to an abstract or overly broad category that is not suitable for object needle grounding tasks, like the sky, sidewalk, slope, etc., please ignore the below instructions and output ``NONE'' only to skip.\\
The output should consist of three lines, separated by a newline:\\
1. A clear scene description, starting with ``Scene: ''.\\
2. The corresponding question based on the scene description, starting with ``Scene Question: ''.\\
3. A concrete object description, starting with ``Object: ''.\\
**Restriction Policies**:\\
- Use the provided object description and subtitle selectively, as they may contain noise.\\
- The scene description should clearly depict the context where the object is in.\\
- The scene description should be a complete sentence.\\
- The scene description should not contain any information that can only be obtained from subtitles.\\
- The question should contain the above scene description and ask when this scene was shown in the video.\\
- The object description should clearly identify the object based on its appearance or context to avoid any ambiguity without referencing bounding boxes. Please do not use the provided noisy object description directly. The data may be correct or may contain errors, so you need to observe the frames yourself and think critically to determine if the data is verifiable.\\
- The object description should be a noun phrase.\\
- Do not use ``red bounding box'', ``image'', or ``frame'' in the answer.\\
- Do not include any additional information or explanations.\\
- Use proper noun phrases when necessary instead of giving general descriptions.\\
Video Frames: \\
\bottomrule
\end{tabular}
}
\label{tab:data_generation_prompts}
\end{table*}

\begin{table*}[t!]
\centering
\caption{Prompts for needle grounding QA self-verification.}
\setlength{\tabcolsep}{0.7mm}
{
\begin{tabular}{p{\textwidth}}
\hline
\toprule
\textbf{Needle Grounding QA Self-Verification}: \\
\midrule
Please help me validate whether the object description uniquely refers to the object in the red bounding box.\\
Given an object description, a scene description, and some frames from a video, where the first \{positive frame number\} frames contain the same object marked with a red bounding box. The remaining \{negative frame number\} frames are sampled from the same video.\\
Object Description: \{object description\}\\
Scene Description: \{scene description\}\\
The output should consist of two lines, each separated by a newline:\\
 - In the first line, you should perform the reasoning and inference process for the validation.\\
 - In the second line, you should state if the data is valid based on the previous reasoning process, using only 'yes' or 'no'.\\
Validation Requirements:\\
 1. Ensure the object description can unambiguously identify the object in the red bounding box based on its appearance or context.\\
 2. Ensure The object description does not refer to any objects in the remaining frames.\\
 3. Ensure the scene description is consistent with the frames with the red bounding box.\\
 4. Ensure the scene description does not contain any information that can only be obtained from the subtitles.\\
 5. Ensure the remaining frames do not match the scene description.\\
Important Notes:\\
 - Please do not have any prior assumptions about the data quality. The data may be correct or may contain errors, so you need to observe the video yourself and think critically to determine if the data is verifiable.\\
 - Make sure the output should be no more than 50 words.\\
The first \{positive frame number\} frames which contain the object in the red bounding box: $<$interleaved$>$\\
The remaining \{negative frame number\} frames:  \\
\bottomrule
\end{tabular}
}
\label{tab:data_generation_prompts2}
\end{table*}

\begin{table*}[t!]
\centering
\caption{Templates for generating object needle grounding questions.}
\setlength{\tabcolsep}{0.7mm}
{
\begin{tabular}{p{0.7\textwidth}}
\hline
\toprule
\textbf{Templates}: \\
\midrule
When does \{object description\} show in the video?\\
At what time does \{object description\} appear in the video?\\
When does \{object description\} appear in the video?\\
When can we see \{object description\} in the video?\\
At what time in the video is \{object description\} visible?\\
When is \{object description\} displayed in the video?\\
What is the timing of \{object description\}'s appearance in the video?\\
When in the video does \{object description\} make an appearance?\\
Can you pinpoint when \{object description\} is shown in the video?\\
When does the video feature \{object description\}? \\
\bottomrule
\end{tabular}
}
\label{tab:object_localization_qa_template}
\end{table*}

\end{document}